\documentclass[iicol,sn-basic]{sn-jnl}

\usepackage{graphicx}%
\usepackage{multirow}%
\usepackage{amsmath,amssymb,amsfonts}%
\usepackage{amsthm}%
\usepackage{mathrsfs}%
\usepackage[title]{appendix}%
\usepackage{xcolor}%
\usepackage{textcomp}%
\usepackage{manyfoot}%
\usepackage{booktabs}%
\usepackage{algorithm}%
\usepackage{algorithmicx}%
\usepackage{algpseudocode}%
\usepackage{listings}%
\usepackage{cleveref}%
\usepackage{hyperref}%
\usepackage{graphicx}%
\usepackage{adjustbox}%
\usepackage{threeparttable}%

\theoremstyle{thmstyleone}%
\newtheorem{theorem}{Theorem}
\newtheorem{proposition}{Proposition}

\newtheorem{remark}{Remark}%

\usepackage{arydshln}

\begin{document}

\title[Article Title]{Geometry-aware Distance Measure for Diverse Hierarchical Structures in Hyperbolic Spaces}

\author[1]{Pengxiang Li}\email{pengxiangli@bit.edu.cn}

\author[1,2]{Yuwei Wu}\email{wuyuwei@bit.edu.cn}

\author*[1]{Zhi Gao}\email{zhi.gao@bit.edu.cn}

\author[1]{Xiaomeng Fan}\email{fanxiaomeng@bit.edu.cn}

\author[1]{Wei Wu}\email{weiwu@bit.edu.cn}

\author*[2]{\\Zhipeng Lu}\email{zhipeng.lu@hotmail.com}

\author[2,1]{Yunde Jia}\email{jiayunde@bit.edu.cn}

\author[3]{and Mehrtash Harandi}\email{mehrtash.harandi@monash.edu}


\affil[1]{\orgdiv{Beijing Laboratory of Intelligent
Information Technology}, \orgname{School of Computer Science, Beijing
Institute of Technology (BIT)}, 
\orgaddress{ \city{Beijing}, \postcode{100081}, \country{P.R. China}}}

\affil[2]{\orgdiv{Guangdong Laboratory of Machine
Perception and Intelligent Computing}, 
\orgname{Shenzhen MSU-BIT University}, 
\orgaddress{ \city{Shenzhen}, \postcode{518172}, \country{P.R. China}}}

\affil[3]{\orgdiv{Department of Electrical and Computer
Systems Eng}, \\
\orgname{Monash University}, \orgaddress{ \city{Clayton VIC}, \postcode{3800}, \country{Australia}}}


\abstract{Learning in hyperbolic spaces has attracted increasing attention due to its superior ability to model hierarchical structures of data. Most existing hyperbolic learning methods use fixed distance measures for all data, assuming a uniform hierarchy across all data points.
However, real-world hierarchical structures exhibit significant diversity, making this assumption overly restrictive. In this paper, we propose a geometry-aware distance measure in hyperbolic spaces, which dynamically adapts to varying hierarchical structures. Our approach derives the distance measure by generating tailored projections and curvatures for each pair of data points, effectively mapping them to an appropriate hyperbolic space. We introduce a revised low-rank decomposition scheme and a hard-pair mining mechanism to mitigate the computational cost of pair-wise distance computation without compromising accuracy. We present an upper bound on the low-rank approximation error using Talagrand's concentration inequality, ensuring theoretical robustness. Extensive experiments on standard image classification (MNIST, CIFAR-10/100), hierarchical classification (5-level CIFAR-100), and few-shot learning tasks (mini-ImageNet, tiered-ImageNet) demonstrate the effectiveness of our method. Our approach consistently outperforms learning methods that use fixed distance measures, with notable improvements on few-shot learning tasks, where it achieves over 5\% gains on mini-ImageNet. The results reveal that adaptive distance measures better capture diverse hierarchical structures, with visualization showing clearer class boundaries and improved prototype separation in hyperbolic spaces.}



\maketitle %

\section{Introduction}\label{sec1}
   Hyperbolic space is defined as a smooth Riemannian manifold with constant negative curvature. Compared with the commonly used Euclidean space, a notable property of hyperbolic spaces lies in the exponential growth of a ball's volume relative to its radius, mirroring the exponential increase in the volume of hierarchical data with depth. Such a property enables the hyperbolic space to serve as a continuous analogue to trees~\citep{pmlr-v80-sala18a,balazevic2019multi}, allowing for modeling hierarchical data with minimal distortion~\citep{sarkar2011low}. The use of hyperbolic spaces for data embeddings has shown superiority over Euclidean space across many applications such as classification~\citep{gao2021curvature,ijcai2022p517,gao2022hyperbolic,atigh2025simzsl}, clustering~\citep{ijcai2022p0451,Lin_2023_ICCV,liu2025hyperbolic}, retrieval~\citep{Ermolov_2022_CVPR}, detection~\citep{hong2024_tnnls,Li_2024_CVPR,gonzalez2025hyperbolic}, segmentation~\citep{hsu2021learning,Atigh_2022_CVPR}, multi-modal learning~\citep{Hong_2023_ICCV,Long_2023_ICCV,mandica2025hyperbolic}, and 3D vision~\citep{montanaro2022rethinking, Leng_2023_ICCV,lin2023hyperbolic,li2025deep}.

    Most existing hyperbolic learning methods employ a fixed distance measure, \textit{i.e.}, the geodesic distance ~\citep{khrulkov2020hyperbolic, Ermolov_2022_CVPR, lin2023hyperbolic}, to assess the similarities between data points.
    Employing a fixed distance measure implicitly assumes a uniform hierarchical structure across all data points~\citep{behrstock2019hierarchically}.
    However, this assumption does not always hold in real-world scenarios, as the hierarchical structures among data points are diverse and complex~\citep{pal2013discovering}. 
    Thus, using a fixed distance measure in real-world scenarios may cause data distortion for diverse hierarchical structures, leading to suboptimal performance.
 
\begin{figure*}[htbp]
    \centering
    \includegraphics[width=0.95\linewidth]{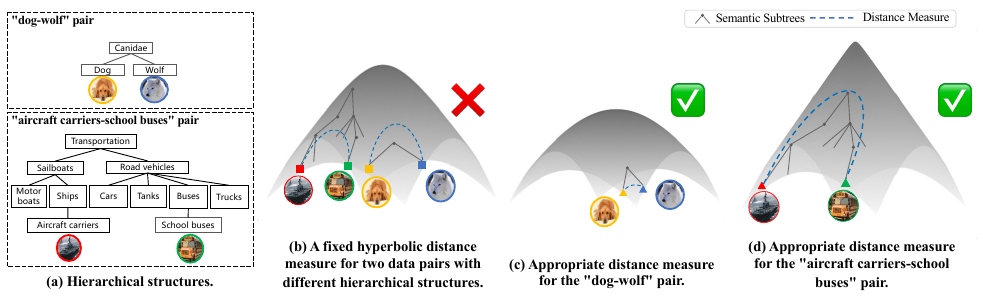}
    \centering
    \caption{Modeling data with diverse hierarchical structures will benefit from adaptive distance measures. 
    (a) The ``dog-wolf" data pair has a simple semantic hierarchical structure, while the ``aircraft carrier-school bus" data pair has a more complex structure. (b) A fixed distance measure fails to fit the hierarchical structures of the two data pairs simultaneously. (c) Flatter distance measures are suitable for the simple structure. (d) Steeper distance measures are suitable for the complex structure.
    The blue dashed lines represent distance measures in hyperbolic space. The solid lines represent the semantic subtrees. 
    }

    \label{fig:abstract}
\end{figure*} 
    
    For example, as shown in \Cref{fig:abstract} (a), the hierarchical structures of data pairs can be reflected via connecting paths between two nodes on the corresponding graph/tree~\citep{behrstock2019hierarchically}. 
    The ``dog-wolf'' data pair has a simple semantic hierarchical structure, while the ``aircraft carrier–school bus'' data pair has a more complex structure. 
    In this case, a fixed distance measure fails to accurately represent the two different hierarchical relationships, that is, the computed distances do not pass through the ancestors of the data pairs, leading to a misalignment between distances and hierarchical structures. As detailed in \Cref{fig:abstract} (b), the ``dog-wolf'' pair and ``aircraft carrier–school bus'' pair have different similarities with different hierarchical structures, but exhibit similar distance values in the embedding space if a fixed distance measure is used. Thus, employing adaptive distance measures for data with diverse hierarchical structures seems to be a natural choice, which could take the complexity of the hierarchical structures into consideration and dynamically fit the diverse hierarchical structures. As shown in Figure \ref{fig:abstract} (c) and (d),  the adaptive distance measure produces shorter paths for the ``dog-wolf'' pair and longer paths for the ``aircraft carrier–school bus'' pair, better reflecting their respective hierarchical complexities.

    In this paper, we propose a geometry-aware distance measure that dynamically adapts to diverse hierarchical structures in hyperbolic spaces.
    The main idea is to learn to generate adaptive projections and curvatures for each pair of samples in hyperbolic spaces, thereby conforming to the hierarchical relationships between data points.
    To achieve this, we design a curvature generator to produce adaptive curvatures for different data pairs, and a projection matrix generator to map data pairs from the original hyperbolic space into adaptive hyperbolic spaces characterized by the newly produced curvature.
    By applying adaptive projections and curvatures to different sample pairs, we obtain our distance measures.

    Two challenges need to be solved in learning to generate geometry-aware distance measures in hyperbolic space. (1) Given that unique projections and curvatures are required for every data pair, reducing the computational cost becomes a critical consideration. 
    (2) As a variety of hierarchical structures are encountered during training, the training stability for pair-wise learning is not easily maintained.

  To address the first challenge, we introduce a low-rank decomposition scheme and a hard-pair mining mechanism. The former reduces the computational complexity of the projection matrix via low-rank approximation. The latter eliminates easy samples by only generating geometry-aware distances for the remaining challenging data pairs.
We theoretically prove that the approximation errors of low-rank decomposition are bounded with high probability.
For the second challenge, we show that incorporating residual connections into the projection matrix during the generation process can effectively maintain training stability.
    
    Experimental results across standard image classification (MNIST, CIFAR-10/100), hierarchical classification (5-level CIFAR-100), and few-shot learning tasks (mini-ImageNet, tiered-ImageNet) confirm the effectiveness of our method in refining hyperbolic learning through geometry-aware distance measures. The experiments reveal that adaptive distance measures effectively capture diverse hierarchical structures, as evidenced by improved class separation in visualizations and better alignment between learned representations and underlying data hierarchies.

    The primary contributions of our work can be summarized as follows:
    \begin{itemize}
        \item We propose a geometry-aware distance measure in hyperbolic spaces that can handle diverse hierarchical structures by learning to produce adaptive curvatures and hyperbolic projections.
        \item We introduce a low-rank decomposition scheme and a hard-pair mining mechanism to significantly reduce computational costs without compromising accuracy. 
        \item  We theoretically prove that the approximation errors of our low-rank decomposition scheme are bounded, guaranteeing its effectiveness in hyperbolic spaces.
    \end{itemize}

\section{Related Work}

\subsection{Hyperbolic Geometry}

Hyperbolic geometry, a non-Euclidean geometry defined by constant negative curvature, permits multiple parallel lines through any given point relative to a reference line. Its intrinsic tree-like structure naturally models hierarchical data, motivating the development of hyperbolic learning algorithms for applications such as few-shot learning~\citep{hamzaoui2024hyperbolic,yu2025large} and 3D vision~\citep{lin2023hyperbolic}.

Early methods typically employ conventional Euclidean neural networks to extract features, followed by a mapping of the resulting embeddings into hyperbolic space~\citep{khrulkov2020hyperbolic,atigh2022hyperbolic,shin2022robust,wang2025learning,leng2025dual}. For example, \citep{liu2020hyperbolic} used a deep convolutional network to obtain image features, which are then projected into hyperbolic space via an exponential map to yield hierarchy-aware representations that enhance zero-shot recognition. However, this two-stage approach can introduce data distortions due to the reliance on a Euclidean backbone. To mitigate these issues, recent works have focused on designing neural networks that operate directly in hyperbolic spaces. Notable examples include hyperbolic convolutional networks~\citep{bdeir2024fully}, hyperbolic residual networks~\citep{van2023poincare}, and hyperbolic binary neural networks~\citep{chen2024_tnnls}.

Parallel efforts in hyperbolic learning algorithms have further improved performance by refining hyperbolic embeddings through optimized metrics~\citep{yan2021unsupervised,liu2020hyperbolic,lucas_2020_hyperml,fan2023horospherical} or exploring adaptive curvatures to better capture varying data structures~\citep{gao2021curvature,Hong_2023_ICCV,yang2022khgcn}. Moreover, novel distance measures have been proposed, including those based on the Lorentzian model~\citep{law2019lorentzian}, tree-level \(l_1\) distances~\citep{yim2023fitting}, point-to-set metrics~\citep{ma2022adaptive}, hyperbolic-aware margins~\citep{yang_hicf_2022,xu2023hyperbolic}, and the integration of vision transformers with hyperbolic embeddings~\citep{ermolov2022hyperbolic}.
In contrast to methods that rely on fixed distance measures, our approach learns geometry-aware hyperbolic distance metrics tailored to the unique hierarchical structure of each data pair.

\subsection{Curvature Learning}
Existing methods~\citep{gu2019learning,khrulkov2020hyperbolic, Ermolov_2022_CVPR,yang_hicf_2022,fang2023poincare} have demonstrated that selecting an appropriate curvature is critical for developing effective learning methods in hyperbolic spaces, as curvature directly influences the quality of learned representations. Research in this area has progressed, ranging from the use of a constant curvature~\citep{bachmann2020constant} in graph neural networks to advancements in curvature learning methods~\citep{yang2023kappahgcn,yang_hicf_2022,gao2021curvature,gao2022pami}. Gu et al.~\citep{gu2019learning} introduced a product manifold approach that embeds hierarchical data by combining multiple spaces with heterogeneous curvatures. Building upon these foundations, our model extends this idea by dynamically adjusting the curvature for pair-wise distance measures. Unlike prior methods, our approach adapts curvature for each pair during inference, leveraging their specific characteristics to enhance the distance measure.

\subsection{Adaptive Deep Metric Learning}
Adaptive metric learning plays a crucial role in machine learning by refining embeddings or distance measures to accommodate data diversity~\citep{li2021adaptive, Li_2019_CVPR,yoon2020xtarnet,liu2021episode}.
Adaptive embeddings are constructed using flexible prototypes~\citep{li2021adaptive}, tailored discriminative features~\citep{Li_2019_CVPR,yoon2020xtarnet}, or episode-specific learning strategies~\citep{liu2021episode}. Similarly, adaptive distance measures encompass task-specific metric spaces~\citep{NEURIPS2018_66808e32, Qiao_2019_ICCV,10040944} and dynamically optimize classifiers using subspaces~\citep{Simon_2020_CVPR}. Recent advancements, such as neighborhood-adaptive metric learning~\citep{9405473,Li_Li_Xie_Zhang_2022}, have further improved adaptability. However, these methods predominantly operate in Euclidean space, limiting their effectiveness for data with complex hierarchical structures. In contrast, we propose learning adaptive distance measures in hyperbolic space through geometry adaptation, leveraging the intrinsic hierarchical property of data for improved structural alignment.

\section{Mathematical Preliminaries}

In the following sections, $\mathbb{R}^n$ denotes an $n$-dimensional Euclidean space and $\|\cdot\|$ denotes the Euclidean norm. 
Vectors are denoted by bold lower-case letters, such as $\boldsymbol{x}$ and $\boldsymbol{y}$. The matrices are denoted by bold upper-case letters, such as $\boldsymbol{M}$. 
The Poincaré ball model of an $n$-dimensional hyperbolic space with curvature $c~(c<0)$ is defined as a Riemannian manifold $\mathbb{B}^n_c = \{ \boldsymbol{x}\in \mathbb{R}^n: -c\|\boldsymbol{x}\| < 1,c<0 \} $, an open ball with radius $1/\sqrt{|c|}$.
We choose the Poincaré ball model in this paper since it provides more flexible operations and gradient-based optimization compared to other hyperbolic models.
The tangent space at $\boldsymbol{x} \in \mathbb{B}^n_c$, a Euclidean space, is denoted by $T_{\boldsymbol{x}}\mathbb{B}^n_c$. 
We use the Möbius gyrovector space~\citep{ungar2001hyperbolic}, which provides operations for hyperbolic learning. Several commonly used operations are summarized as follows.


\noindent\textbf{Addition.} For $\boldsymbol{x}, \boldsymbol{y} \in \mathbb{B}_c^n$, the Möbius addition is defined as
\begin{equation}
\boldsymbol{x} \oplus_c \boldsymbol{y}=\frac{\left(1-2 c\langle\boldsymbol{x}, \boldsymbol{y}\rangle-c\|\boldsymbol{y}\|^2\right) \boldsymbol{x}+\left(1+c\|\boldsymbol{x}\|^2\right) \boldsymbol{y}}{1-2 c\langle\boldsymbol{x}, \boldsymbol{y}\rangle+c^2\|\boldsymbol{x}\|^2\|\boldsymbol{y}\|^2}
.\end{equation}

\noindent\textbf{Geodesic Distance.} The geodesic distance $d_c(\cdot, \cdot)$ between two points $\boldsymbol{x}, \boldsymbol{y} \in \mathbb{B}_c^n$ is given by
\begin{equation}
d_c(\boldsymbol{x}, \boldsymbol{y})=\frac{2}{\sqrt{|c|}} \operatorname{arctanh}\left(\sqrt{|c|}\left\|-\boldsymbol{x} \oplus_c \boldsymbol{y}\right\|\right).
\label{eq:dist_0}
\end{equation} 

\noindent\textbf{M\"{o}bius Matrix-vector Multiplication.} In hyperbolic spaces, the M\"{o}bius matrix-vector multiplication $\otimes_c$ is defined for a matrix $\boldsymbol{M} \in \mathbb{R}^{n\times n}$
and a vector $\boldsymbol{x} \in \mathbb{B}^{n}_{c}$,
\begin{equation}
    \scalebox{0.98}{ $
        \boldsymbol{M} \otimes_c \boldsymbol{x} = \frac{1}{\sqrt{|c|}}\mathrm{tanh}(\frac{\|\boldsymbol{M}\boldsymbol{x}\|}{\|\boldsymbol{x}\|}\mathrm{arctanh}(\sqrt{|c|}\|\boldsymbol{x}\|))\frac{\boldsymbol{M}\boldsymbol{x}}{\|\boldsymbol{M}\boldsymbol{x}\|}.
        $}
    \label{eq:bmm}
\end{equation}

\noindent\textbf{Exponential Map.} The exponential map $\mathrm{expm}^c_{\boldsymbol{x}}(\boldsymbol{v})$ projects a vector $\boldsymbol{v}$ from the tangent space $T_{\boldsymbol{x}}\mathbb{B}^n_c$ to the Poincar\'e ball $\mathbb{B}^n_c$,
\begin{equation}
\scalebox{0.98}{ $
\operatorname{expm}_{\boldsymbol{x}}^c(\boldsymbol{v})=\boldsymbol{x} \oplus_c\left(\tanh \left(\sqrt{|c|} \frac{\lambda_{\boldsymbol{x}}^c\|\boldsymbol{v}\|}{2}\right) \frac{\boldsymbol{v}}{\sqrt{|c|}\|\boldsymbol{v}\|}\right).
 $}
\end{equation}

\noindent\textbf{Logarithmic Map.} The logarithmic map $\operatorname{logm}_{\boldsymbol{x}}^c (\boldsymbol{y} )$ maps a vector $\boldsymbol{y}$ from the Poincar\'e ball $\mathbb{B}^n_c$ to the tangent space  $T_{\boldsymbol{x}}\mathbb{B}^n_c$,  
\begin{equation}
\scalebox{0.88}{ $
\operatorname{logm}_{\boldsymbol{x}}^c(\boldsymbol{y})=\frac{2}{\sqrt{|c|} \lambda_{\boldsymbol{x}}^c} \mathrm{arctanh}\left(\sqrt{|c|}\left\|-\boldsymbol{x} \oplus_c \boldsymbol{y}\right\|\right) \frac{-\boldsymbol{x} \oplus_c \boldsymbol{y}}{\left\|-\boldsymbol{x} \oplus_c \boldsymbol{y}\right\|}.
$}
\end{equation}

\noindent\textbf{Hyperbolic Averaging.} We use $\mathrm{Einstein}$ $\mathrm{mid}$-$\mathrm{point}$ as the counterpart of Euclidean averaging in hyperbolic space. The $\mathrm{Einstein}$ $\mathrm{mid}$-$\mathrm{point}$ has the simplest form in the Klein model $\mathbb{K}$. 
For $\{\boldsymbol{x}_1, \ldots, \boldsymbol{x}_N\}, \boldsymbol{x}_i \in \mathbb{B}^n_c$, we first map $\boldsymbol{x}_i$ from $\mathbb{B}^n_c$ to $\mathbb{K}$, then perform averaging in Klein model, and finally map the mean in $\mathbb{K}$ back to $\mathbb{B}^n_c$ to obtain the Poincar\'e mean:
\begin{equation}
\scalebox{0.98}{ $
     \boldsymbol{u}_{i} =  \frac{2 \boldsymbol{x}_{i}}{1+c\left\|\boldsymbol{x}_{i}\right\|^{2}}, \ \
        \overline{\boldsymbol{u}} =  \frac{\sum_{i=1}^{N} \gamma_{i} \boldsymbol{u}_{i}}{\sum_{i=1}^{N} \gamma_{i}}, \ \
        \overline{\boldsymbol{x}} =  \frac{\overline{\boldsymbol{u}}}{1+\sqrt{1-c\|\overline{\boldsymbol{u}}\|^{2}}},
        $}
\label{eq:midpoint}
\end{equation}
where $\boldsymbol{u}_{i} \in \mathbb{K}$, $\overline{\boldsymbol{u}}$ is the mean in $\mathbb{K}$, $\overline{\boldsymbol{x}}$ is the mean in $\mathbb{B}^n_c$, and $\gamma_i=\frac{1}{\sqrt{1-c\left\|\boldsymbol{x}_i\right\|^2}}$ is the Lorentz factor.

\noindent\textbf{$\delta$-hyperbolicity.} The $\mathrm{Gromov}$ $\mathrm{\delta}$-$\mathrm{hyperbolicity}$ \citep{gromov1987hyperbolic} is a measure of how closely the hidden structure of data resembles a hyperbolic space. A lower value of $\delta$ indicates a higher degree of intrinsic hyperbolicity in the data. The $\mathrm{Gromov}$ $\mathrm{\delta}$-$\mathrm{hyperbolicity}$ is computed as follows. First, we start from the $\mathrm{Gromov \ product}$ for $\boldsymbol{x,y,z} \in \mathbb{X}$, denoted as
\begin{equation}
    (\boldsymbol{y,z})_x = \frac{1}{2} \bigg(d\left(\boldsymbol{x,y}\right) + d\left(\boldsymbol{x,z}\right) - d\left(\boldsymbol{y,z}\right)\bigg),
\end{equation}
where $\mathbb{X}$ is an arbitrary space endowed with the distance function $d$. Following  \citep{fournier2015computing}, we compute the pair-wise Gromov product of all the data, and the results of all pairs are denoted as a matrix $\boldsymbol{A}$. Then $\mathrm{\delta}$-$\mathrm{hyperbolicity}$ is computed by
\begin{equation}
    \delta = (\max_k\min\{\boldsymbol{A}_{ik}, \boldsymbol{A}_{kj}\}) - \boldsymbol{A}.    
\end{equation} Relative $\mathrm{\delta}$-$\mathrm{hyperbolicity}$ is computed by $\delta_{rel} = \frac{2\delta(\mathbb{X})}{\mathrm{diam}(\mathbb{X})} \in [0, 1]$, where $\mathrm{diam} (\mathbb{X})$ denotes the set diameter (maximum pair-wise distance). Values of $\delta_{rel}$ closer to 0 indicate a stronger hyperbolicity of a dataset. 

 \begin{figure*}[htbp]
     \centering
     \includegraphics[width=1\linewidth]{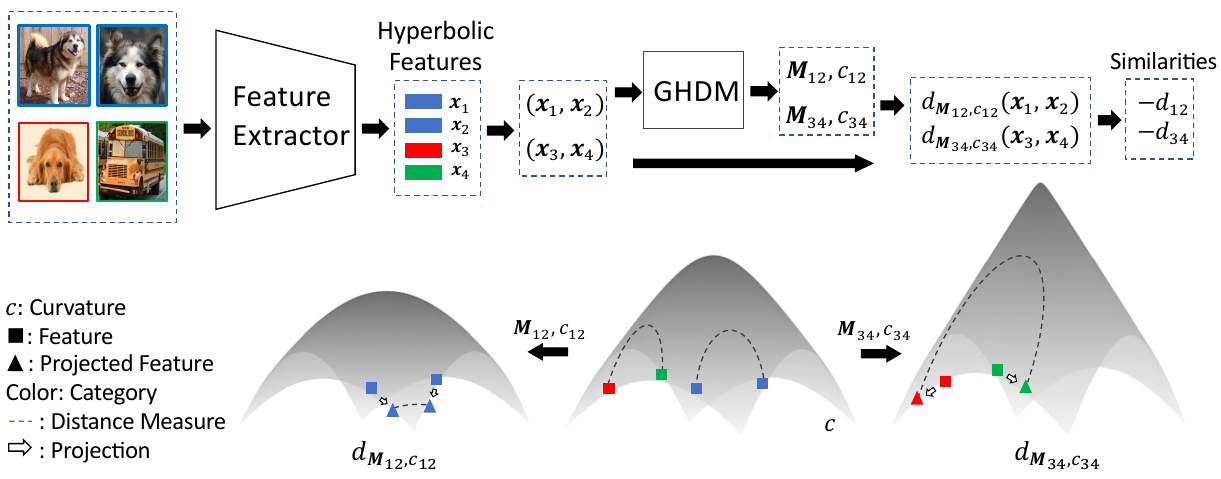}
     \caption{Overview of the proposed method. Two data pairs (one positive and one negative) are encoded with the feature extractor, while the color of the embeddings indicates the class of each data point. ``GHDM'' indicates the geometry-aware hyperbolic distance measure that generates the distance measures $d_{\boldsymbol{M}_{12}, c_{12}}$ and $d_{\boldsymbol{M}_{34}, c_{34}}$ according to the given data pair. Under the adapted distance measure with adaptive projection and curvature, the positive samples are pulled closer, while the negative samples are pushed further away. 
     }
     \label{fig:pipeline}
 \end{figure*}

\section{Method}
\subsection{Analysis}
In this section, we demonstrate that effective hierarchical structure modeling requires geometry-aware distance measures, including curvature adaptation and feature projections. 

\vspace{-10pt}
\begin{proposition}
In a hyperbolic space $\mathbb{B}^n_c$, larger $|c|$ produces steeper geodesics $d_c(\boldsymbol{x}_i, \boldsymbol{x}_j)$ that can better represent complex hierarchical structures.
\end{proposition}
\vspace{-15pt}
\begin{proof}



We define the complexity of hierarchical structures in hyperbolic space. Let \(\mathbf{a}\) denote the common ancestor of two data points \(\boldsymbol{x}_1,\boldsymbol{x}_2\in\mathbb{B}^{n}_c\),
the complexity can be defined as
\[
C(\boldsymbol{x}_1,\boldsymbol{x}_2)=P(\boldsymbol{x}_1\rightarrow \mathbf{a})+P(\boldsymbol{x}_2\rightarrow \mathbf{a}),
\]
where \(P(\cdot\rightarrow \mathbf{a})\) denotes the connectivity to \(\mathbf{a}\) measured by the graph distance \citep{balbuena1996distance}. This complexity is equivalent to counting the number of hierarchical levels traversed along the unique tree path connecting \(\boldsymbol{x}_1\) and \(\boldsymbol{x}_2\) through their common ancestor. This shows that complex hierarchical structures have more hierarchical levels, and thus cause steep distances (\emph{i.e., the distance curve close to the origin}).

\begin{figure}
    \centering
    \includegraphics[width=0.7\linewidth]{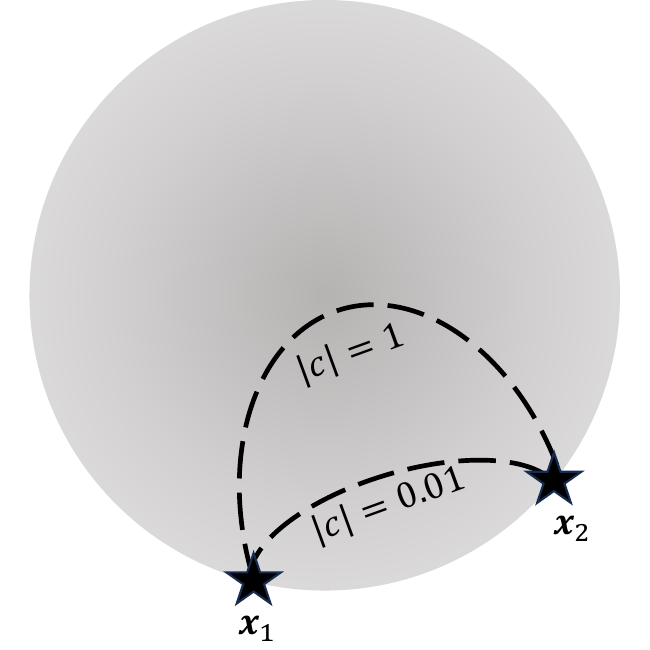}
    \caption{Relationship between curvature and geodesic steepness in the Poincaré ball.}
    \label{fig:curvature}
\end{figure}

In hyperbolic space, the geodesic distance between the pair \((\boldsymbol{x}_1,  \boldsymbol{x}_2)\) is given by Eq.~(\ref{eq:dist_0}):
$$
d_c(\boldsymbol{x}_1, \boldsymbol{x}_2)=\frac{2}{\sqrt{|c|}} \operatorname{arctanh}\left(\sqrt{|c|}\left\|-\boldsymbol{x}_1 \oplus_c \boldsymbol{x}_2\right\|\right),
$$ 
where larger $|c|$ corresponds to steeper geodesics ~\citep{nickel2017poincare,krioukov2010hyperbolic}, as shown in Figure~\ref{fig:curvature}., which accommodate higher complexity \(C(\boldsymbol{x}_1,\boldsymbol{x}_2)\) by providing appropriate geometric scale to distinguish complex hierarchical relationships.
\end{proof}

\vspace{-10pt}
\begin{proposition} 
Hierarchical data with diverse structures requires adaptive projections for data points beyond uniform geodesic distances to capture varying local structural patterns.
\end{proposition}
\vspace{-15pt}
\begin{proof}

Consider a dataset $\mathcal{D}$ with multiple subtrees $\{\mathcal{S}_1, \mathcal{S}_2, \ldots, \mathcal{S}_K\}$. 
Different subtrees exhibit distinct structural patterns, such as varying hierarchical complexities, intra-class differences, inter-class differences, feature importance, and noise levels. 
As shown in Euclidean metric learning methods~\citep{kulis2012metric}, learning projections for data points can incorporate the structural patterns, consistently outperforming fixed metrics.
We argue it is also reasonable in hyperbolic spaces.
For example, given data pairs with complex hierarchical structures, their deep hierarchical levels require projections, through which the data pairs have larger value differences to adequately capture these deep levels.
Therefore, proper projections lead to effective hierarchical modeling.
\end{proof}


These analyses establish that hierarchical structure modeling in hyperbolic spaces requires adaptive distance measures with curvature adaptation and embedding projection.
Using both of them motivates the development of geometry-aware distance measures for real-world hierarchical data.

\subsection{Formulation}
\begin{figure*}
     \centering
     \includegraphics[width=0.8\linewidth]{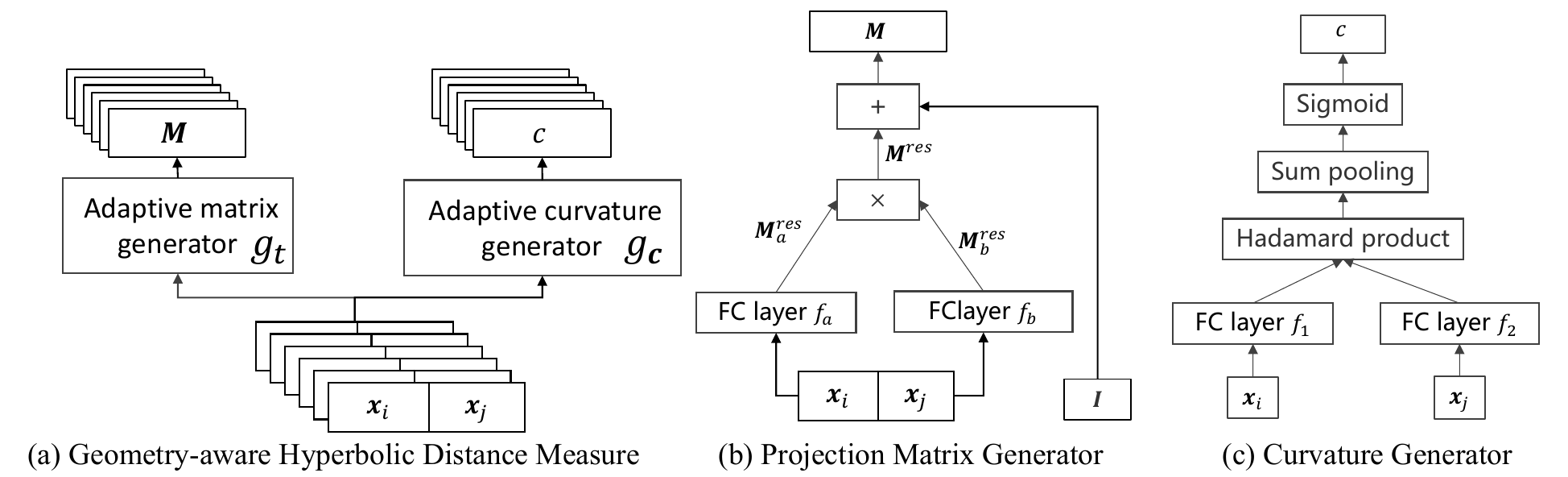}
     \caption{(a) Architecture of the geometry-aware distance measure including: (b) the projection matrix generator $g_t(\cdot)$ and (c) the curvature generator $g_c(\cdot)$. ${\boldsymbol{x}}$ are the feature points.}
     \label{fig:subnet}
 \end{figure*}

The schematic overview of our method is illustrated in Figure~\ref{fig:pipeline}. We start by employing a feature extractor to obtain hyperbolic representations ${\boldsymbol{x}}$ of the data (images in our implementation).
After extracting hyperbolic features, our method produces an adaptive projection matrix $\boldsymbol{M}_{ij}$ and curvature $c_{ij}$ for one pair of hyperbolic features $(\boldsymbol{x}_{i}, \boldsymbol{x}_j)$. 
For ease of exposition, we will use $\boldsymbol{M}$ to denote $\boldsymbol{M}_{ij}$ and use $c$ to refer to $c_{ij}$.
In this case, we define the geometry-aware distance between $(\boldsymbol{x}_{i}, \boldsymbol{x}_j)$ as
\begin{equation}
    d_{\boldsymbol{M}, c}(\boldsymbol{x}_{i}, \boldsymbol{x}_j) = d_{c}(\boldsymbol{M} \otimes_{c} \boldsymbol{x}_{i}, \boldsymbol{M} \otimes_{c}\boldsymbol{x}_j), 
    \label{eq:dist_1}
\end{equation}
where $d_{c}(\cdot,\cdot)$ is the Poincar\'e geodesic distance function (Eq.~(\ref{eq:dist_0})).
$\boldsymbol{M}$ is produced by a matrix generator $g_t (\cdot)$: $ \boldsymbol{M} = g_t(\boldsymbol{x}_i, \boldsymbol{x}_j)$, and $c$ is generated by a curvature generator $g_c (\cdot)$: $ c = g_c(\boldsymbol{x}_i, \boldsymbol{x}_j)$. As depicted in Figure~\ref{fig:subnet}, we show their architectures.

The goal of our method is to train the matrix generator $g_t (\cdot)$ and the curvature generator $g_c (\cdot)$, through which positive pairs are closer and negative pairs are pushed farther apart in Eq.~\eqref{eq:dist_1}.
In practical applications, generating projection matrices and curvatures for all pairs would result in significant computational costs. To reduce the computation overhead, we introduce a hard-pair mining mechanism to select hard pairs $\mathcal{H} = \mathrm{HPM}(D_{s})$, where  $\mathcal{H}$ contains hard cases, $D_{s}$ is the support set, and $\mathrm{HPM}(\cdot)$ denotes the hard-pair mining mechanism. Our method only generates projection matrices and curvatures for pairs in $\mathcal{H}$.


 \subsection{Projection Matrix Generator}

The projection matrix generator $g_t(\cdot)$ takes feature pairs $(\boldsymbol{x}_{i}, \boldsymbol{x}_j)$ as input and outputs a transformation matrix $\boldsymbol{M}$ for computing geometry-aware distances. Its architecture is shown in Figure~\ref{fig:subnet} (b).

\paragraph{Residual Learning Framework.} 
To ensure training stability, we avoid directly generating the full transformation matrix $\boldsymbol{M}$. Instead, we adopt a residual learning approach by learning the difference $\boldsymbol{M}^{res}$ between our geometry-aware distance measure (Eq.~(\ref{eq:dist_1})) and the standard Poincaré distance (Eq.~(\ref{eq:dist_0})). The projection matrix is then computed as
\begin{equation}
\boldsymbol{M} = \mathbf{I} + \boldsymbol{M}^{res}.
\end{equation}

This formulation naturally recovers the standard Poincaré geodesic distance when $\boldsymbol{M}^{res} = \mathbf{0}$, providing a principled way to extend the base distance measure.

\paragraph{Low-Rank Decomposition.} 
When the embedding dimension $n$ is large, generating and operating on the full matrix $\boldsymbol{M}^{res} \in \mathbb{R}^{n \times n}$ becomes computationally expensive. To address this challenge, we decompose $\boldsymbol{M}^{res}$ into a product of two low-rank matrices:
\begin{equation}
\boldsymbol{M}^{res} = \boldsymbol{M}^{res}_{a} {\boldsymbol{M}^{res}_{b}}^{\top}.
\end{equation}
where $\boldsymbol{M}^{res}_{a}, \boldsymbol{M}^{res}_{b} \in \mathbb{R}^{n \times k}$ with $k \ll n$. These matrices are generated by fully connected networks:
\begin{align}
\boldsymbol{M}^{res}_{a} &= f_{a}(\boldsymbol{x}_i, \boldsymbol{x}_j), \\
\boldsymbol{M}^{res}_{b} &= f_{b}(\boldsymbol{x}_i, \boldsymbol{x}_j).
\end{align}

The final projection matrix becomes
\begin{equation}
\boldsymbol{M} = \mathbf{I} + \boldsymbol{M}^{res}_{a} {\boldsymbol{M}^{res}_{b}}^{\top}.
\label{eq:hmg_2}
\end{equation}

This reduces computational complexity from $O(n^2)$ to $O(nk)$ while maintaining approximation quality.

\paragraph{Theoretical Justification.} 
We provide theoretical guarantees for our low-rank approximation based on the polynomial partitioning theory~\citep{guth2015onthe,talagrand1995concentration}. Let $\boldsymbol{M} = \mathbf{I} + \boldsymbol{M}^{res}_{a}{\boldsymbol{M}^{res}_{b}}^{\top}$ denote our low-rank approximation and $\boldsymbol{M}' = \mathbf{I} + \boldsymbol{M}^{res}$ denote the full-rank target matrix.
\vspace{-15pt}
\begin{theorem}
\label{theo:main}
Suppose the hyperbolic features $\{\boldsymbol{x}\}$ is continuous. For any $\boldsymbol{x} \in \mathbb{B}^{n}_c$, the approximation error $|\boldsymbol{M}\otimes_c\boldsymbol{x} - \boldsymbol{M'}\otimes_c\boldsymbol{x}|$ is bounded by $O(n^{-\beta})$ with the probability at least $1 - O(e^{-\sqrt{k}})$, where $\beta \in (0,1)$ is a small constant and $k$ is the rank of $\boldsymbol{M}^{res}_{a}$ and ${\boldsymbol{M}^{res}_{b}}$.
\end{theorem}
\vspace{-30pt}
\begin{remark}
Theorem~\ref{theo:main} establishes that our low-rank approximation maintains high accuracy with high probability, justifying the computational efficiency gains from the decomposition.
\end{remark}
\vspace{-20pt}
\begin{proof}
We decompose the approximation error in Möbius matrix-vector multiplication $\otimes_c$ into scalar and directional components, then bound each one using classical techniques.

For one vectors $\boldsymbol{x}$ with $\|\boldsymbol{x}\| \leq c < 1$, the Möbius matrix-vector multiplication can be written as
\[
\boldsymbol{M}\otimes \boldsymbol{x} = a \cdot b,
\]
where $a = \tanh\left(\frac{\|\boldsymbol{M}\boldsymbol{x}\|}{\|\boldsymbol{x}\|}\operatorname{arctanh}(\|\boldsymbol{x}\|)\right)$ (scalar component) and $b = \frac{\boldsymbol{M}\boldsymbol{x}}{\|\boldsymbol{M}\boldsymbol{x}\|}$ (directional component).

The approximation error satisfies
\[
\|\boldsymbol{M}\otimes\boldsymbol{x} - \boldsymbol{M}'\otimes\boldsymbol{x}\| \leq |a - a'| + \|b - b'\|.
\]

\noindent\textbf{Scalar Component:} Define Rayleigh quotients $\delta = \frac{\|\boldsymbol{M}\boldsymbol{x}\|}{\|\boldsymbol{x}\|}$ and $\delta' = \frac{\|\boldsymbol{M}'\boldsymbol{x}\|}{\|\boldsymbol{x}\|}$. By the mean value theorem ~\citep{vadhan2013uniform}, we have
\[
|a-a'| \leq u \cdot |\delta-\delta'|,
\]
where $u = \operatorname{arctanh}(\|\boldsymbol{x}\|)$ is bounded. Using the Eckart-Young-Mirsky theorem~\citep{eckart1936approximation,mirsky1960symmetric}, we bound $|\delta-\delta'|$ through optimal subspace projection. Thus, the scalar component $|a - a'|$ is bounded.

\noindent\textbf{Directional Component:} $b$ and $b'$ are unit vectors, and the difference $\|b - b'\|$ is controlled using randomized analysis on the sphere $S^{n-1}$. By partitioning data into local regions with approximately Gaussian distributions and applying Talagrand's concentration inequality~\citep{talagrand1995concentration}, we obtain $\|b-b'\| < n^{-\beta}$ with high probability,  where $\beta \in (0,1)$ characterizes the convergence rate of approximation error with increasing embedding dimension $n$.

Combining both components and assuming bounded Rayleigh quotients~\citep{hwang2004cauchy}, the overall error satisfies
\[
\|\boldsymbol{M}\otimes\boldsymbol{x} - \boldsymbol{M}'\otimes\boldsymbol{x}\| = O(n^{-\beta}),
\]
with probability at least $1 - O(e^{-\sqrt{k}})$, where $k$ is the rank of our low-rank decomposition that controls the trade-off between computational efficiency and approximation quality (chosen as $k \ll n$ in practice). 
The larger the rank $k$ is, the higher the probability becomes, and the better the approximation quality becomes, but at the cost of increased computational complexity that scales linearly with $k$. The exponential probability bound in $\sqrt{k}$ ensures strong theoretical guarantees even for moderate rank values.
\end{proof}

\subsection{Curvature Generator}

In the curvature generator $g_c(\cdot)$, we employ factorized bilinear pooling~\citep{gao2020revisiting} to generate adaptive curvatures for sample pairs $(\boldsymbol{x}_i, \boldsymbol{x}_j)$. We leverage second-order feature interactions to capture the geometric warping properties essential for hyperbolic space representation. Its architecture is shown in Figure~\ref{fig:subnet} (c).

The process consists of three steps. First, we transform the input samples using separate fully connected layers: $\boldsymbol{x}^\prime_{i} = f_1(\boldsymbol{x}_i)$ and $\boldsymbol{x}^\prime_{j} = f_2(\boldsymbol{x}_j)$. Second, we compute their Hadamard product: $\boldsymbol{B} = \boldsymbol{x}^\prime_i \circ \boldsymbol{x}^\prime_j$. Finally, we convert $\boldsymbol{B}$ to curvature $c$ through sum pooling followed by the sigmoid activation $\sigma$ to ensure $c \in [0,1]$:
\begin{equation}
    c = \sigma(\mathrm{sum\_pooling}(\boldsymbol{B})).
    \label{eq:curvature}
\end{equation}

\vspace{-25pt}
\begin{remark}
Factorized bilinear pooling captures second-order feature interactions that are beneficial for modeling hierarchical structures.
These interactions can encode the complex non-linear relationships between features, enabling the model to accurately represent the intrinsic geometric properties and warping characteristics of the hyperbolic space, leading to accurate curvature estimation, as proven in~\citep{gao2021curvature}. But the difference is that the work~\citep{gao2021curvature} produces curvatures for tasks, while we produce curvatures for data pairs.
\end{remark}

\begin{figure*}
\centering
\includegraphics[width=0.75\textwidth]{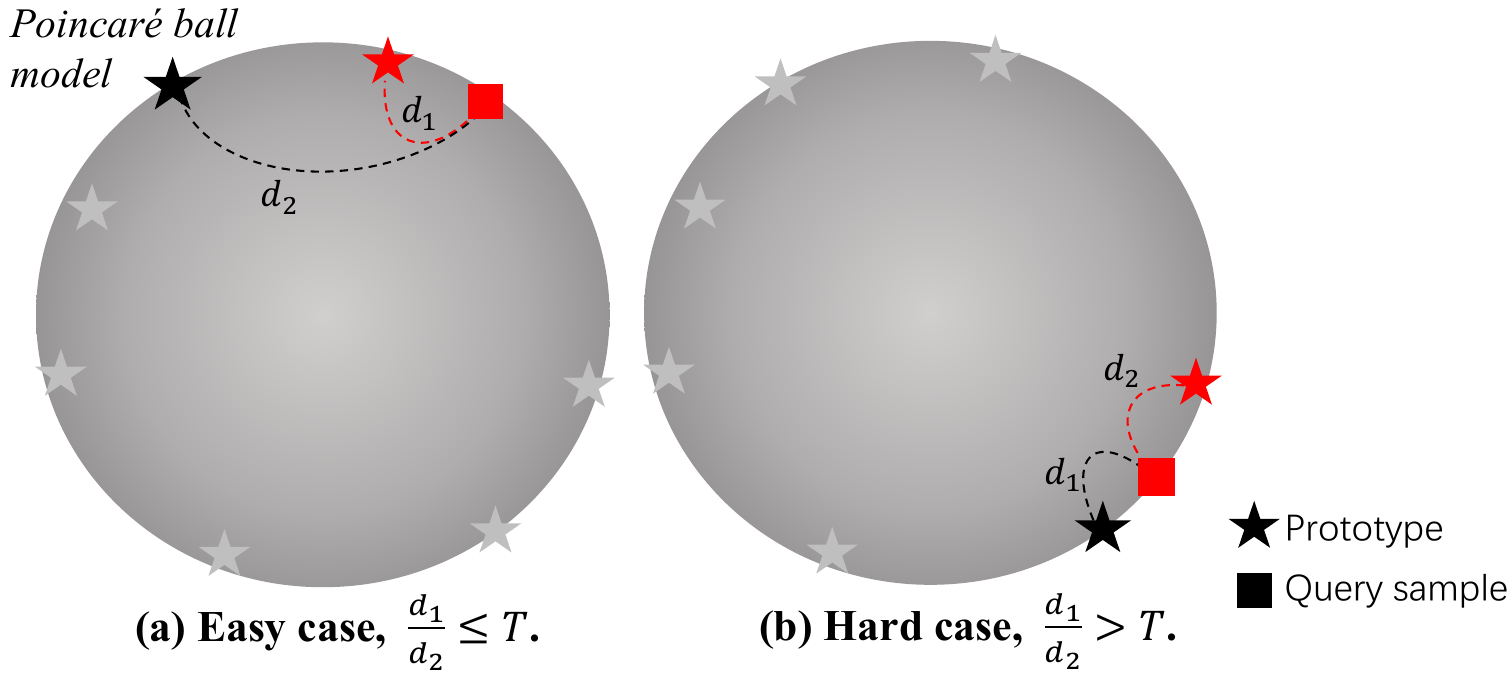}
\caption{The proposed hard-pair mining mechanism in hyperbolic space. The threshold \( T \in [0,1] \) is a margin hyperparameter. Embedding classes are color-coded, and Poincaré geodesic distances are shown as dashed lines. The distances from the query sample to the closest and second closest prototypes are \( d_1 \) and \( d_2 \), respectively. Hard cases, difficult to classify, are defined by \( d_1/d_2 > T \).}
\label{fig:mining}
\end{figure*}

\subsection{Hard-pair Mining in Hyperbolic Space}



Creating geometry-aware distance measures may incur high computational costs. To tackle this, we propose a hard-pair mining mechanism $\mathrm{HPM}(\cdot)$ to identify difficult pairs, $\mathcal{H} = \mathrm{HPM}(D_{s})$, where $\mathcal{H}$ contains hard cases, $D_{s}$ is the support set.
Our approach only generates distance measures and computes loss values for pairs in $\mathcal{H}$.

For the classification task, we compute hyperbolic distances between one query sample and all prototypes in the Poincaré ball model. We then identify the two closest prototypes, with distances $d_1$ (nearest) and $d_2$ (second nearest). 
We compute the ratio $d_1/d_2$ to measure the difficulty of classifying this query sample.
A ratio closer to 1 indicates a harder case for accurate classification. For a feature set $\mathcal{F}_{q}$ of the query set and the prototype set $\mathcal{P}$, the hard-case mining process can be represented as
\begin{equation}
   \mathcal{H} = \{\boldsymbol{x} \mid \boldsymbol{x} \in \mathcal{F}_{q}, \frac{d_c(\boldsymbol{x},\boldsymbol{p}_{1})}{d_c(\boldsymbol{x},\boldsymbol{p}_{2})} > T\},
   \label{eq:mining}
\end{equation}
where $d_c(\cdot, \cdot)$ is the Poincar\'e geodesic distance function in Eq.~(\ref{eq:dist_0}),
$\boldsymbol{p}_1 = \displaystyle\operatorname{argmin}_{\boldsymbol{p} \in \mathcal{P}} d_c(\boldsymbol{x}, \boldsymbol{p})$, and $\boldsymbol{p}_2 = \displaystyle\operatorname{argmin}_{\boldsymbol{p} \in \mathcal{P} - \boldsymbol{p}_1} d_c(\boldsymbol{x}, \boldsymbol{p})$.   
$T \in [0,1]$  is a margin hyperparameter. Figure~\ref{fig:mining} shows the hard-pair mining mechanism applied to the Poincaré ball model.

\subsection{Training Process}
The goal of this work is to learn the projection matrix generator $g_t$ and the curvature generator $g_c$. 
During the training process, we partition the training set $\mathcal{D}$ into the support set $\mathcal{D}_{s}$ and the query set $\mathcal{D}_{q}$. 
We extract features $\mathcal{F}_{s}$ and $\mathcal{F}_{q}$ from $\mathcal{D}_{s}$ and $\mathcal{D}_{q}$, respectively, using the backbone network with the exponential map $\mathrm{expm}^c_{\boldsymbol{x}}(\boldsymbol{\cdot})$.
We denote the set of prototypes as $\mathcal{P}$ that are the Einstein mid-points computed from $\mathcal{F}_{s}$. 
The hard cases $\mathcal{H}$ are selected from $\mathcal{F}_q$ via hard-pair mining in Eq.~(\ref{eq:mining}). Then we generate the adaptive matrices $\boldsymbol{M}$ and curvatures $c$ using $g_t$ and $g_c$ via Eq.~(\ref{eq:hmg_2}) and Eq.~(\ref{eq:curvature}) for data pairs in $\mathcal{H} \times \mathcal{P}$, and compute geometry-aware distances as the logits via Eq.~(\ref{eq:dist_1}) and Eq. ~(\ref{eq:logit}),
\begin{equation}
    \textbf{logits} =  \mathrm{softmax}(- 
    d_{\boldsymbol{M}, c}(\mathcal{H},\mathcal{P})).
    \label{eq:logit}
\end{equation}
We update $g_t$ and $g_c$ by minimizing the cross entropy loss of $\mathcal{D}_q$, as shown in Eq. ~(\ref{eq:cross-entropy}). 

\begin{equation}
    Loss = - \sum_{i=1}^{C} \textbf{y}_{i} \log(\textbf{logits}),
    \label{eq:cross-entropy}
\end{equation}
where $C$ is the number of the class and $y_i$ is the ground-truth label.
The pseudo-code of training is summarized in Algorithm \ref{al:training}.


\begin{algorithm}
  \small
  \caption{Training process of our method}
  \label{al:training}
  \begin{algorithmic}[1]
    \Require Training set $\mathcal{D}$
    \Ensure Updated metric generators $g_t$ and $g_c$
    \While{not converged}
      \State Randomly sample a support set $\mathcal{D}_s$ and a query set $\mathcal{D}_q$ from $\mathcal{D}$.
      \State Extract features $\mathcal{F}_s$ and $\mathcal{F}_q$ from $\mathcal{D}_s$ and $\mathcal{D}_q$.
      \State Compute Einstein-midpoint prototypes $\mathcal{P}$ from $\mathcal{F}_s$.
      \State Select hard cases $\mathcal{H}$ from $\mathcal{F}_q$ via Eq.~(\ref{eq:mining}).
      \State Generate adaptive matrices and curvatures via $g_t, g_c$ (Eqs.~(\ref{eq:hmg_2}), (\ref{eq:curvature})) for data pairs in $\mathcal{H} \times \mathcal{P}$, then compute logits (Eqs.~(\ref{eq:dist_1}), (\ref{eq:logit})).
      \State Compute cross-entropy loss via Eq.~(\ref{eq:cross-entropy}) and update $g_t, g_c$.
    \EndWhile
  \end{algorithmic}
\end{algorithm}

\subsection{Complexity Analysis}

Low-rank distance measures reduce computational costs in generating projection matrices. Our method produces two low-rank matrices ($n \times k$) with a cost of $O(n^2k)$, compared to $O(n^3)$ for directly generating an $n \times n$ matrix, where $k \ll n$.

The time complexities of our method's components are $O(pq)$ for hard-pair mining, $O(n^2)$ for curvature generation, and $O(n^2k)$ for projection matrix generation. Here, $p$ is the number of classes, $q$ is the number of queries, $n$ is the dimensionality, and $k$ is the rank. The overall time complexity is $O(n^2(k+1)+pq) \approx O(n^2(k+1))$, as $p$ and $q$ are much smaller than $n$.

\begin{table}[]
  \centering
  \small
    \caption{The relative delta $\delta_{rel}$ values calculated for used datasets. Lower delta $\delta_{rel}$ values denotes higher hyperbolicity.}
    \begin{tabular}{lcc}
      \hline
      \textbf{Dataset} & \textbf{Encoder} & $\boldsymbol{\delta}_{rel}$ \\ \hline
      CIFAR10          & Wide-Res 28$\times$2     & 0.354 \\
      CIFAR100         & Wide-Res 28$\times$2     & 0.280 \\
      mini-ImageNet    & ResNet-12                & 0.328 \\
      tiered-ImageNet  & ResNet-12                & 0.228 \\ \hline
    \end{tabular}
    \label{tab:delta}
\end{table}

\section{Experiment}
We evaluate the proposed method on standard classification, hierarchical classification, and few-shot learning tasks. 
Similar to existing methods ~\citep{gao2022hyperbolic,khrulkov2020hyperbolic,Ermolov_2022_CVPR}, we use backbone networks with the exponential map as the feature extractor. Then we apply the hard-pair mining mechanism to select the hard cases, and finally generate the distance measures for the hard cases. We set the threshold as $0.96$ in the hard-pair mining mechanism. The rank $k$ is set to $16$ for all settings. 


\begin{table}[h]
  \centering
  \footnotesize
    \caption{Accuracy (\%) comparisons with existing hyperbolic learning methods on the MNIST, CIFAR10, and CIFAR100 datasets. }
    \begin{tabular}{lccc}
      \hline
      \textbf{Method}      & \textbf{MNIST} & \textbf{CIFAR10} & \textbf{CIFAR100} \\ \hline
      Hyp-Optim            & 94.42          & 88.82            & 72.26             \\
      HNN++                & 95.01          & 91.22            & 73.65             \\
      Hyp-ProtoNet         & 93.53          & 93.30            & 73.83             \\
      Ours                 & \textbf{96.56} & \textbf{94.75}   & \textbf{75.61}    \\ \hline
    \end{tabular}
    \label{tab:std_clf}
\end{table}

\subsection{Standard Classification}
\label{sec:5.1}
\subsubsection{Dataset}

We conduct experiments on three popular datasets, namely MNIST\citep{mnist}, CIFAR10\citep{cifar}, and CIFAR100\citep{cifar}.
MNIST contains 10 classes with 60000 training images and 10000 testing images. Each image has a resolution of $28 \times 28$, and the numerical pixel values are in greyscale.  
The CIFAR-10 dataset consists of 60000 color images in 10 classes, each class having 6000 images with a size of $32 \times 32$. 
CIFAR-100 has 100 classes containing 600  $32 \times 32$ color images each, with 500 training and 100 testing images per class.

\subsubsection{$\delta$-hyperbolicity}
For image datasets, we measure the $\delta$-hyperbolicity based on $l_2$ distances between the features produced by our feature extractors. Values of $\delta_{rel}$ closer to 0 indicate a stronger hyperbolicity of a dataset~\citep{khrulkov2020hyperbolic}.
Results are averaged across 1000 subsamples of 20000 data points. The values of $\delta_{rel}$ on the image datasets we used are shown in \Cref{tab:delta}. As can be seen from \Cref{tab:delta}, these image datasets all have a clear hierarchical structure (the \(\delta_{rel}\) of these datasets is close to 0).

\subsubsection{Results}
We classify query samples by calculating distances between these prototypes and the features of query samples. Table~\ref{tab:std_clf} compares our method with existing hyperbolic learning methods. On MNIST, CIFAR10, and CIFAR100, our method improves by 3.03\%, 1.45\%, and 1.78\% over Hyp-ProtoNet~\citep{khrulkov2020hyperbolic}, and by 2.14\%, 5.93\%, and 3.35\% over Hyp-Optim~\citep{ganea2018hyperbolic}. Compared to HNN++~\citep{shimizu2020hyperbolic}, our method achieves 1.55\%, 3.53\%, and 1.9\% higher accuracy. These results demonstrate that our adaptive distance measures outperform existing hyperbolic learning methods by better matching inherent hierarchical structures.

\begin{table*}[htbp]
  \centering
  \small
    \caption{Hierarchical accuracy (\%) of the fixed distance measure and our method on the CIFAR-100 dataset. Levels 0 to 4 (coarse-to-fine) represent test results at different levels of annotation. For the `fixed' distance measure, the curvature is set to $0.5$.}
    \begin{tabular}{lccccc}
      \hline
      \textbf{Method}        & \textbf{Level 0} & \textbf{Level 1} & \textbf{Level 2} & \textbf{Level 3} & \textbf{Level 4} \\ \hline
      ResNet50+fixed         & 95.62            & 90.65            & 88.68            & 86.30            & 78.49            \\
      ResNet50+ours          & \textbf{96.50}   & \textbf{91.88}   & \textbf{90.22}   & \textbf{88.11}   & \textbf{81.19}   \\ \hline
      ResNet101+fixed        & 95.95            & 91.51            & 90.08            & 87.87            & 80.97            \\
      ResNet101+ours         & \textbf{97.88}   & \textbf{93.68}   & \textbf{92.27}   & \textbf{90.13}   & \textbf{83.44}   \\ \hline
    \end{tabular}
    \label{tab:hie}
\end{table*}

\begin{table*}[htbp]
\centering
\caption{Accuracy (\%) comparisons with popular few-shot learning methods on the mini-ImageNet and tiered-ImageNet datasets. ‘Optim’ and ‘Metric’ mean the optimization-based and metric-based few-shot learning methods, respectively.  ‘Euc’ and ‘Hyp’ mean the methods are performed in Euclidean space and Hyperbolic space, respectively. `*' indicates that results use ResNet-18 \citep{he2016deep} as the backbone, while the others use ResNet-12 \citep{he2016deep}.}
\resizebox{0.96\linewidth}{!}
{
\begin{tabular}{c|c|c|cc|cc}
\hline
\tiny
\multirow{2}{*}{\textbf{Method}}          & \multirow{2}{*}{\textbf{Space}} & \multirow{2}{*}{\textbf{Category}} & \multicolumn{2}{c|}{\textbf{min-ImageNet}}    & \multicolumn{2}{c}{\textbf{tiered-ImageNet}}  \\ 
                                          &                                 &                                    & 1-shot                & 5-shot                & 1-shot                & 5-shot                \\ \hline
MAML \citep{finn2017model_maml}             & Euc                             & Optim                              & 51.03 ± 0.50          & 68.26 ± 0.47          & 58.58 ± 0.49          & 71.24 ± 0.43          \\
L2F \citep{baik2020learning_l2f}            & Euc                             & Optim                              & 57.48 ± 0.49          & 74.68 ± 0.43          & 63.94 ± 0.84          & 77.61 ± 0.41          \\
MeTAL \citep{baik2021meta_metal}            & Euc                             & Optim                              & 59.64 ± 0.38          & 76.20 ± 0.19          & 63.89 ± 0.43          & 80.14 ± 0.40          \\
Meta-AdaM \citep{sun2023metaadam}            & Euc                             & Optim                              & 59.89 ± 0.49         & 77.92 ± 0.43          & 65.31 ± 0.48          & 85.24 ± 0.35          \\
\hline
ProtoNet \citep{snell2017proto}             & Euc                             & Fixed  Metric                      & 56.52 ± 0.45          & 74.28 ± 0.20          & 53.51 ± 0.89          & 72.69 ± 0.74          \\
DSN \citep{Simon_2020_CVPR}                 & Euc                             & Fixed Metric                       & 62.64 ± 0.66          & 78.83 ± 0.45          & 66.22 ± 0.75          & 82.79 ± 0.48          \\
LMPNet \citep{huang2021local} & Euc                             & Fixed Metric                       & 62.74 ± 0.11          & 80.23 ± 0.52          & 70.21 ± 0.15          & 79.45 ± 0.17          \\
TADAM \citep{NEURIPS2018_66808e32}          & Euc                             & Adaptive Metric                    & 58.50 ± 0.30           & 76.70 ± 0.30           & -                     &    -                   \\
XtarNet \citep{yoon2020xtarnet}             & Euc                             & Adaptive  Metric                   & 55.28 ± 0.33          & 66.86 ± 0.31          & \, 61.37 ± 0.36*      & \, 69.58 ± 0.32*      \\
\hline\hline 
Hyp-Kernel \citep{fang2021kernel}  & Hyp                             & Optim                              & \,61.04 ±  0.21*  & \,77.33 ± 0.15*           &  \,57.78 ± 0.23*         & \,76.48 ± 0.18*       \\
C-HNN \citep{guo2022clipped}            & Hyp                             & Optim                              & 53.01 ±  0.22 & 72.66 ± 0.15          & -          &  -          \\
CurAML \citep{gao2022pami}             & Hyp                             & Optim                              & 63.13 ± 0.41 & 81.04 ± 0.39          & 68.46 ± 0.56          & 83.84 ± 0.40          \\
Hyp-ProtoNet \citep{khrulkov2020hyperbolic} & Hyp                             & Fixed  Metric                      & \, 59.47 ± 0.20*      & \, 76.84 ± 0.14*      & -                     & -                     \\ \hline
Ours                                      & Hyp                             & Adaptive  Metric                   & \textbf{64.75 ± 0.20}          & \textbf{81.89 ± 0.15} & \textbf{72.59 ± 0.22} & \textbf{86.14 ± 0.16} \\ \hline
\end{tabular}
}

\label{tab:few_shot}
\end{table*}

\subsection{Hierarchical Classification}

\subsubsection{Dataset}
\label{appendix:hie}
Using the CIFAR100 dataset and its 5-level hierarchical annotations from \citep{wang2023transhp}, we compare our adaptive distance measure with a fixed one, using Resnet50 and Resnet101.
The category IDs of the 5 hierarchical levels are as follows:
 \begin{itemize}
     \item  Level 4 (100 categories): The original fine labels in CIFAR-100.
     \item   Level 3 (20 categories): The original coarse labels in CIFAR-100.
     \item  Level 2 to Level 0: Constructed based on the 20 coarse labels (Level 3), provided by \citep{wang2023transhp}. Specifically:
     \begin{itemize}
        \item    Level 2 with ten categories: ([0-1]), ([2-17]), ([3-4]), ([5-6]), ([12-16]), ([8-11]), ([14-15]), ([9-10]), ([7-13]), ([18-19]).
        \item Level 1 with five categories: ([0-1-12-16]), ([2-17-3-4]), ([5-6-9-10]), ([8-11-18-19]), ([7-13-14-15]).
        \item Level 0 with two categories: ([0-1-7-8-11-12-13-14-15-16]) and ([2-3-4-5-6-9-10-17-18-19]).
     \end{itemize}
         
 \end{itemize}

We report our accuracies on the 5 hierarchical levels using Resnet50 and Resnet101, as shown in \Cref{tab:hie}. Compared with using a fixed distance measure (Eq.~(\ref{eq:dist_0}), denoted as `fixed', where we tune the curvature and set it as $0.5$ to achieve its best performance), our adaptive distance measure has better performance on all hierarchical levels, indicating that classes belonging to the same parent node are closely grouped after our projection. The results demonstrate that our model can effectively capture the implicit hierarchical structure within the data. 

\begin{figure}[htbp]
     \centering
     \includegraphics[width=1.0\linewidth]{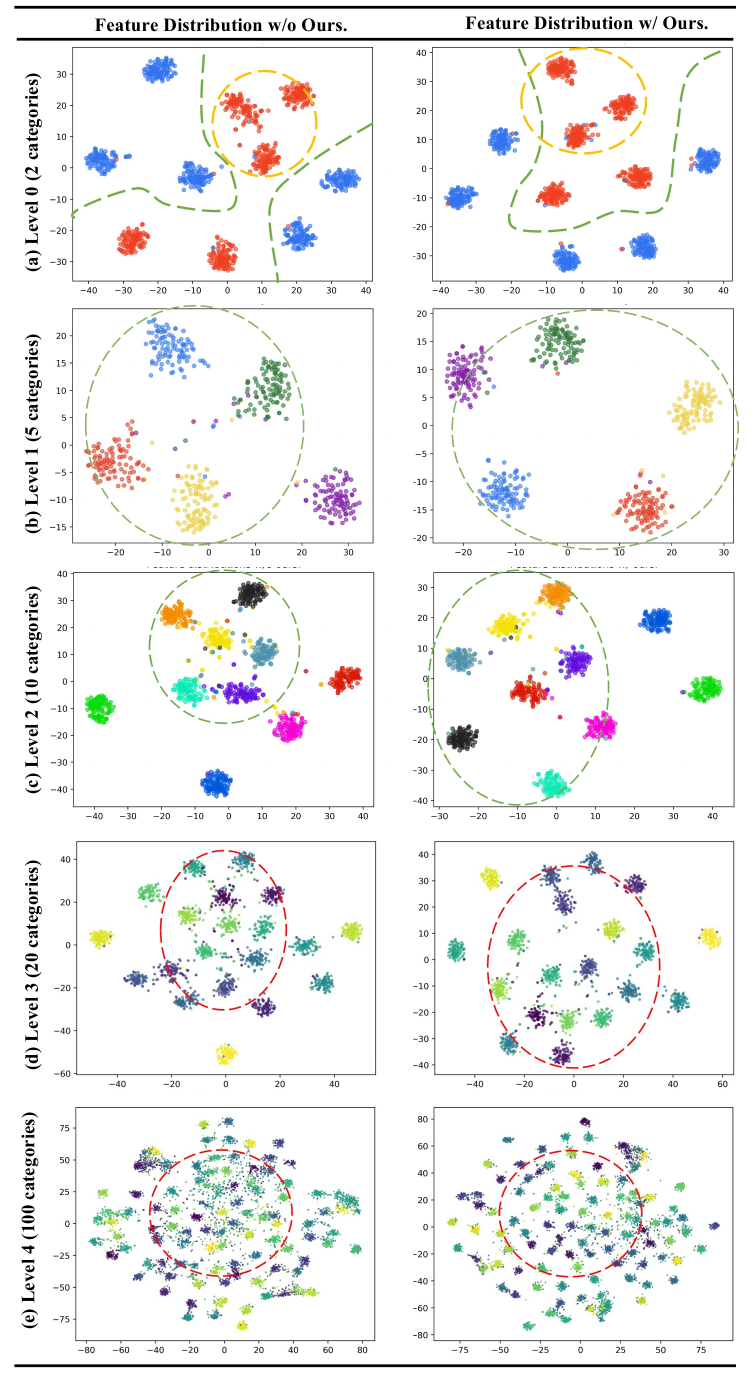}
     \caption{Feature distribution of the 5-level classification on the CIFAR 100 dataset via T-SNE. For levels 1-4, we randomly sampled one fine category (out of 100) from each coarse category. For level 0, we sampled five fine categories from each coarse category. The colors indicate the categories. 
     }
     \label{fig:hie}
 \end{figure}

We visualize the distribution of our hyperbolic embeddings at the 5 hierarchical levels, as shown in Figure~\ref{fig:hie}. We observe that our method leads to clearer boundaries and larger inter-class distances among different categories on the 5 hierarchical levels, showing that our method can better capture hierarchical structures again.

\subsection{Few-shot Learning}
\label{sec:5.2}
We conduct experiments on two popular few-shot learning datasets: mini-ImageNet \citep{vinyals2016matching} and tiered-ImageNet \citep{ren18fewshotssl}.
We compare our method with the hyperbolic methods, including the metric-based Hyp-ProtoNet \citep{khrulkov2020hyperbolic} and the optimization-based Hyp-Kernel \citep{fang2021kernel}, C-HNN \citep{guo2022clipped}, and CurAML\citep{gao2022pami}, as shown in Table~\ref{tab:few_shot}.
Note that Hyp-ProtoNet \citep{khrulkov2020hyperbolic} is a fixed metric-based hyperbolic few-shot learning method. Compared with it, our method is 5.28\% and 5.05\% higher than it on the 1-shot and 5-shot settings, respectively, suggesting that our method generates better distance measures for matching the inherent hierarchical structures of data.
We also compare our method with the popular Euclidean optimization-based methods\citep{finn2017model_maml, baik2020learning_l2f,baik2021meta_metal,gao2021curvature,sun2023metaadam} and Euclidean metric-based methods\citep{snell2017proto, vinyals2016matching,lee2019meta,lu2021tailoring,Simon_2020_CVPR,NEURIPS2018_66808e32,li2020boosting,yoon2020xtarnet,khrulkov2020hyperbolic}. Our method improves their performance on both the 1-shot and the 5-shot tasks. Compared with the fixed Euclidean metric-based methods\citep{snell2017proto,Simon_2020_CVPR,huang2021local}, our method brings more than 1\% improvements on the 1-shot task and 2\% on the 5-shot task. Compared with the adaptive metric-based methods in the Euclidean space, such as TADAM \citep{NEURIPS2018_66808e32} and XtarNet \citep{yoon2020xtarnet}, our method exceeds them in both 1-shot and 5-shot accuracy. The main reason is that performing metric learning in the hyperbolic space preserves the hierarchical structures of data and avoids undesirable data distortion.

\begin{table*}[]
 \centering
\footnotesize
  \caption{5-shot accuracy(\%) and and 95 \% confidence interval on mini-ImageNet and tiered-ImageNet dataset. The hard-pair mining is deactivated.}

    \begin{tabular}{llcc}
    \hline
    \textbf{ID} &\textbf{Metric}&\textbf{mini-ImageNet} &\textbf{tiered-ImageNet} \\ \hline

    (\romannumeral1)                 & Fixed hyp metric & 81.38 ± 0.20 & 83.94 ± 0.16                        \\
    (\romannumeral2)                 & Ours w/o $g_t$  & 81.40 ± 0.14   & 84.67 ± 0.15                        \\
    (\romannumeral3)                & Ours w/o $g_c$  &81.57 ± 0.14 & 84.85 ± 0.16                        \\
    (\romannumeral4)                 & Ours    &  \textbf{81.80 ± 0.14}      & \textbf{86.10 ± 0.16}                       \\ \hline
    \end{tabular}%

 \label{tab:abla}
\end{table*}

\subsection{Ablation}
\subsubsection{Hyperbolic Distance Measure}
 
We evaluate the effectiveness of different components in our method, the projection matrix generator and the curvature generator, on the mini-ImageNet datasets and the tiered-ImageNet dataset. 
(\romannumeral1)  We evaluate the fixed hyperbolic metric by employing the Poincaré geodesic distance function, as defined in Eq.~(\ref{eq:dist_0}). 
(\romannumeral2) We evaluate only using the adaptive curvature. We deactivate the projection matrices generator $g_t$ and solely utilize the adaptive curvature generator $g_c$ with Eq.~(\ref{eq:dist_0}). 
(\romannumeral3) We evaluate only using the adaptive projection matrices. We activate the $g_t$ but disable the $g_c$, setting the curvature to 0.5. Results are shown in Table~\ref{tab:abla}.

In Table~\ref{tab:abla}, the adaptive curvature generator $g_c$ provides a more discriminative feature space than the fixed curvature space, \textit{i.e.}, (\romannumeral1) \textit{vs.}(\romannumeral2). 
As evidenced in rows (\romannumeral1) and (\romannumeral3) of Table~\ref{tab:abla}, our method benefits from the projection matrix generator $g_t$, and the geometry-aware distance measures match better with the inherent hierarchical structures than the fixed distance measure.

\begin{table*}[h]
\centering
\footnotesize
\caption{5-shot accuracy(\%), memory cost(MB) and time cost(ms) per few shot learning task with different ranks during inference. The memory cost of one inference process with the setting of 5w5s and 15 queries. We disable the hard-pair mining when testing the memory cost. }
    \begin{tabular}{ccccc}
    \hline
     &   \textbf{Rank}  & \textbf{5-shot acc(\%}) & \textbf{Mem(MB)} & \textbf{Time(ms)} \\ \hline
       &  4  & 81.41 ± 0.14 & 98.06 & 5.02 \\ 
       & 8  & 81.53 ± 0.14 & 179.56 & 5.25 \\
     \small{{w/ ours}}  &  16 & 81.80 ± 0.14 & 352.86 & 6.10 \\ 
        &  32 & 81.48 ± 0.14 & 703.10 & 8.67 \\ 
         & 64 & 81.45 ± 0.14 & 1371.64 & 14.26 \\ \hline
     \small{{w/o ours}} &  512 & 81.74 ± 0.14 & 7231.24 & 60.88 \\ \hline
    \end{tabular}
      \label{tab:rank}
\end{table*}

\subsubsection{Rank in Matrix Decomposition}
We further explore the effect of rank in matrix decomposition on the mini-ImageNet dataset. 
Our low-rank decomposition decomposes the original $n \times n$ matrices (for ResNet-12 backbone, $n=512$) into two $n \times k$ matrices multiplication, greatly reducing the computational complexity from $O(n^3)$ to $O(nk^2), k << n$. 
Here, we evaluate the value of $k$ in the range of $[4,8,16,32,64]$, and report the accuracy and memory cost.
As shown in Table~\ref{tab:rank}, the accuracy increases first and then decreases as the rank increases. As the rank increases, we retain more and more information, resulting in an increase in accuracy. However, when the rank becomes too large, excessive information, including errors and noise, may be preserved, which can lead to overfitting and a decrease in accuracy.  
As rank increases, the number of model parameters and the computational cost both increase significantly. As shown in Table~\ref{tab:rank}, when $k = 64$, the total memory consumption is nearly four times that of $k = 16$. 
Considering the trade-off between accuracy and computational cost, this paper selects $k = 16$.
 
\subsubsection{Hard-pair Mining}
\begin{table*}[htbp]
\centering

\caption{Effectiveness of the hard-pair mining. Threshold is the $T = d_{1} / d_{2}$. The `Percentage' represents the proportion of hard cases in the query set. The rest columns represent the 5-shot  accuracy(\%) of the easy cases with Eq.~(\ref{eq:dist_0}), hard cases with Eq.~(\ref{eq:dist_0}), hard cases with our methods, total query set with Eq.~(\ref{eq:dist_0}) and total query set with our method, respectively.}
\resizebox{1\linewidth}{!}
{
\begin{tabular}{c|c|c|cc|cc}
\hline
\textbf{Threshold} & \textbf{Percentage} & \textbf{Easy cases w/ Eq.~(\ref{eq:dist_0})} & \textbf{Hard cases w/ Eq.~(\ref{eq:dist_0})} & \textbf{Hard cases w/ ours} & \textbf{Total w/ Eq.~(\ref{eq:dist_0})} & \textbf{Total w/ ours} \\ \hline
0.1                & 100\%               & -                   & 81.26 ± 0.14                 & \textbf{81.61 ± 0.14 }              & 81.26 ± 0.14            & \textbf{81.61 ± 0.14}          \\
0.8                & 89\%                & 98.48 ± 0.14        & 78.97 ± 0.14                 & \textbf{79.39 ± 0.14 }              & 81.18 ± 0.14            & \textbf{81.54 ± 0.14}          \\
0.9                & 57\%                & 97.07 ± 0.08        & 69.17 ± 0.17                 & \textbf{69.91 ± 0.17 }              & 81.14 ± 0.14            & \textbf{81.53 ± 0.14 }         \\
0.96               & 22\%                & 89.29 ± 0.11        & 53.52 ± 0.24                 & \textbf{55.62 ± 0.25  }             & 81.38 ± 0.14            & \textbf{81.89 ± 0.14 }         \\
0.98               & 14\%                & 86.51 ± 0.13        & 49.40 ± 0.35                 & \textbf{52.41 ± 0.36 }              & 81.27 ± 0.14            & \textbf{81.66 ± 0.14  }        \\
0.99               & 7\%                 & 83.73 ± 0.23        & 45.92 ± 0.51                 & \textbf{50.35 ± 0.53   }            & 81.21 ± 0.23            & \textbf{81.49 ± 0.23 }         \\ \hline
\end{tabular}
}
\label{tab:mining}
\end{table*}

\noindent\textbf{Accuracy.}
We further evaluate the effectiveness of the hard-pair mining mechanism on the mini-ImageNet dataset, as shown in Table~\ref{tab:mining}. We assess the threshold value $T$ in the range of $[0.1, 0.8, 0.9, 0.96, 0.98, 0.99]$ and report 5-shot accuracy using both the fixed distance measure (Eq.~(\ref{eq:dist_0})) and the adaptive distance measure (Eq.~(\ref{eq:dist_1})) for easy cases, hard cases, and the total query set.
The results show that our mechanism effectively selects hard examples. At \( T = 0.9 \), 57\% of samples are classified as hard. As the threshold increases, the proportion of hard cases decreases, reducing computational complexity by focusing on adaptive measures for hard cases. The mechanism also distinguishes effectively between easy and hard cases, with higher accuracy for the former.
Our geometry-aware distance measure significantly improves classification accuracy for hard cases. As difficulty increases, the benefits of our method become more pronounced. For thresholds of 0.96, 0.98, and 0.99, accuracy improvements for hard cases are 2.1\%, 3.01\%, and 4\%, respectively, compared to cases without our method.

 \begin{table}[htbp]
    
    \centering
    \caption{Percentage of hard pairs, 5-shot accuracy(\%) and time per few-shot learning task on the mini-ImageNet dataset. HPM denotes the hard-pair mining, and the threshold $T$ is set as $0.96$.}
    \begin{tabular}{lcc}
    \hline
        & \textbf{w/o HPM} &  \textbf{w/ HPM} \\ \hline
        {Hard-pair percentage(\%)} &     100 &  22 \\
        {5-shot acc(\%)} & 81.80 ± 0.14 & 81.89 ± 0.15 \\
        {Time of HPM(ms)} &  0 & 0.243 \\
        {Total running time(ms)} &  6.10 & 1.52 \\ \hline
 
    \end{tabular}

    \label{tab:hpm}
\end{table}

\noindent\textbf{Efficiency.} The results presented in Table \ref{tab:hpm} indicate that hard-pair mining effectively filters out 79\% of all pairs while requiring only 0.243 ms per few-shot learning task, where we set the threshold $T$ as $0.96$. This leads to a significant decrease in the total run time, from 6.10 ms to just 1.52 ms. This reduction stems from computing distances for only 21\% of pairs selected by the hard-pair mining. Additionally, using hard-pair mining even slightly improves performance (81.80\% \textit{vs.} 81.89\%).

\subsection{Effectiveness of the Residual Connection}
\begin{figure}
    \centering
    \includegraphics[width=1.0\linewidth]{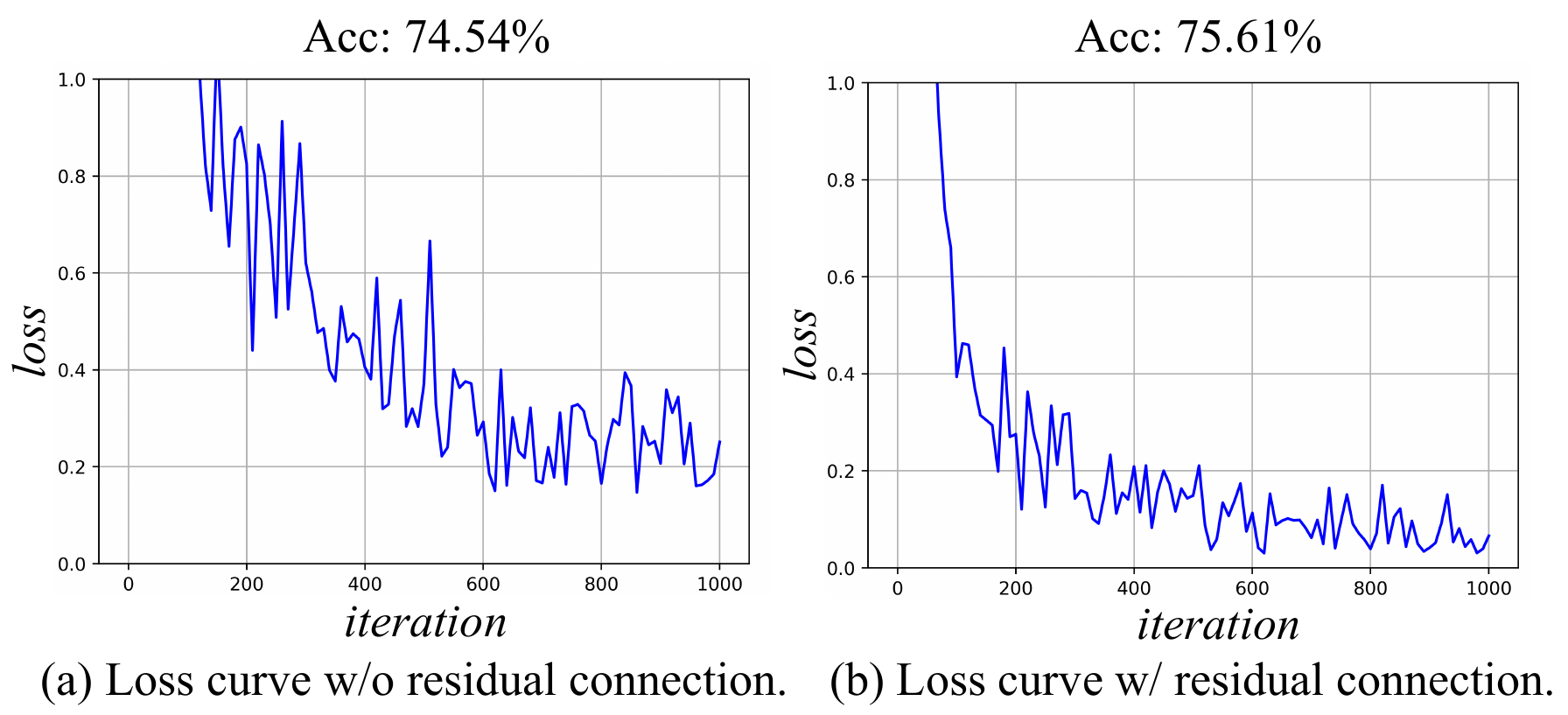}
    \caption{Loss curves on the CIFAR100 dataset. }
    \label{fig:res}
\end{figure}

We conduct experiments about the accuracy and training loss w/ and w/o the residual connection to verify its effectiveness on the CIFAR100 dataset. Using the residual solution brings 1.07\% improvements (w/o res (74.54\%) \textit{vs.} w/ res(75.61\%)).  The loss curves in Figure~\ref{fig:res} show that using the residual connection brings a stable training process with faster convergence and smoother loss curves.

\begin{table*}[htbp]
    \centering
     \small
\caption{Ablations about the symmetry. 5-shot accuracy(\%) and and 95 \% confidence interval on the mini-ImageNet and tiered-ImageNet.}
    \begin{tabular}{lcc}
    \hline
      & \textbf{mini-ImageNet} & \textbf{tiered-ImageNet} \\ \hline
    {Ours} &  81.80 ± 0.14 & 85.22 ± 0.16  \\
    {Swapping} $f_a$ {and} $f_b$ & 81.81 ± 0.14 & 85.20 ± 0.16 \\
    {Swapping} $f_1$ {and} $f_2$ & 81.80 ± 0.14 & 85.22 ± 0.16  \\
    \hline
    \end{tabular}
  
    \label{tab:sym}
\end{table*}

\subsection{Exploitation of Symmetry}
Our distance measure cannot ensure the strict symmetry of input pair $(\boldsymbol{x}, \boldsymbol{y})$, since we do not share the weights of the subnetworks ($f_a$ and $f_b$ in the projection matrix generator $g_t$, $f_1$ and $f_2$ in the curvature generator $g_c$) within the generator, where $f_a$ and $f_1$ are used for $x$, and $f_b$ and $f_2$ are used for $y$. But we do not argue that this is a problem. For a sample pair $(\boldsymbol{x}, \boldsymbol{y})$ in the training process, we randomly sample $x$ and $y$ from the dataset, and the probability of $(\boldsymbol{x}, \boldsymbol{y})$ and $(\boldsymbol{y}, \boldsymbol{x})$ are equal. Thus, if we change $(\boldsymbol{x}, \boldsymbol{y})$ to $(\boldsymbol{y}, \boldsymbol{x})$, and send it to our distance measure, their distances will not change significantly.

To demonstrate this point, we do the ablation by swapping $f_a$ and $f_b$ in the projection matrix generator $g_t$) as well as $f_1$ and $f_2$ in the curvature generator $g_c$, which is equal to swapping the inputs of $g_t$ and $g_c$ separately. As shown in Table \ref{tab:sym}, the impact of swapping the subnetworks in the projection matrix generator and the curvature generator on the performance of our method is negligible. This indicates that $f_a$ and $f_b$, as well as $f_1$ and $f_2$, have learned nearly identical knowledge. Although our method does not ensure symmetry, it avoids situations where swapping $\boldsymbol{x}$ and $\boldsymbol{y}$ in a pair leads to a sharp change in distance, showing the robustness of our method.

\begin{table}
\centering
  \small
    \caption{Ablation study on symmetry. 5-way accuracy (\%) and 95\% confidence interval for contrastive loss  vs. cross-entropy loss  on the mini-ImageNet dataset.}
    \begin{tabular}{lcc}
      \hline
      \textbf{Loss Func} & \textbf{5-way 1-shot} & \textbf{5-way 5-shot} \\ \hline
      Contrastive             & 61.91 ± 0.20           & 79.45 ± 0.14          \\
      Cross-Entropy            & \textbf{64.75 ± 0.20}  & \textbf{81.89 ± 0.15} \\ \hline
    \end{tabular}
    \label{tab:loss}
\end{table}

\subsection{Different Loss Functions}
We have also tried the supervised contrast loss function, specifically, 
\[
\scalebox{0.9}{$L(\boldsymbol{x}_i, \boldsymbol{x}_j, \boldsymbol{x}_k) = \max(0, margin + d(\boldsymbol{x}_i, \boldsymbol{x}_j) - d(\boldsymbol{x}_i, \boldsymbol{x}_k)),$}
\]
where $\boldsymbol{x}_i$ is the anchor sample, $\boldsymbol{x}_j$ is the positive sample, and $\boldsymbol{x}_k$ is the negative sample. 
Our algorithm is capable of converging with this type of loss function as well as shown in \Cref{{tab:loss}}. However, since it equally penalizes all negative samples without considering their hierarchical differences, it may hinder the model's ability to differentiate between less relevant and highly irrelevant samples. Ideally, this loss function would require the true hierarchical structure as ground truth, but such detailed annotations are absent in the current visual datasets. Consequently, we chose direct cross-entropy loss $$L= -\sum_{i=1}^{C} y_i \log(p_i),$$where $y_i$ is the label and $p_i$ is the probability of the prediction, allowing the network to implicitly learn the diverse geometric structures among different pairs.

\begin{table}[htbp]
  \centering
  \small
    \caption{Accuracy (\%) comparison between symmetric design ($\boldsymbol{M}_{a}^{res}{\boldsymbol{M}_{a}^{res}}^\top$) and our design ($\boldsymbol{M}_{a}^{res}{\boldsymbol{M}_{b}^{res}}^\top$) on the CIFAR-10 and CIFAR-100 datasets.}
    \begin{tabular}{lcc}
      \hline
      \textbf{Dataset} & \textbf{Symmetric Form} & \textbf{Ours} \\ \hline
      CIFAR-10         & 88.64                   & \textbf{96.56} \\
      CIFAR-100        & 71.02                   & \textbf{75.61} \\ \hline
    \end{tabular}
    \label{tab:symmatrix}
\end{table}

\subsection{Different Structures of the Projection Matrix Generator}
In the design of the projection matrix generator, we use two separate low-rank residual matrices $\boldsymbol{M}^{res}_{a}  {\boldsymbol{M}^{res}_{b}}^{\top}$ to approximate the projection matrix. We also test the results of using a unified low-rank residual matrix $\boldsymbol{M}^{res}_{a}  {\boldsymbol{M}^{res}_{a}}^{\top}$. Results in \Cref{tab:symmatrix} show that asymmetrical designs outperform symmetrical ones in effectiveness. 

Symmetrical designs are geometrically constrained to axis-aligned transformations~\citep{horn2012matrix} and may not capture the complex hierarchical relationships of real-world scenes. Symmetric matrices can only represent combinations of scaling and reflection along orthogonal axes~\citep{strang2016introduction}, limiting their ability to model arbitrary geometric deformations. In contrast, our proposed geometry-aware distance measure leverages an asymmetric structure, allowing for more diverse geometric transformations and adaptable fitting to complex scene geometries.


 \subsection{Standard Classification on Multiple Backbones}
The experimental results on WideRes-28, ResNet-50, and ResNet-101 using the CIFAR100 dataset are summarized in \Cref{tab:multi_backbone}. Our method demonstrates consistent improvements across all tested backbones, achieving an accuracy increase of 1.78\% on WideRes28$\times$2, 2.7\% on ResNet50, and 2.48\% on ResNet101. These results underscore the effectiveness of the proposed geometry-aware hyperbolic distance measure in enhancing model performance, regardless of the architecture, and confirm its generalizability to diverse network structures.

\begin{table}[htbp]
  \centering
  \small
    \caption{Accuracy (\%) on WideRes-28, ResNet-50, and ResNet-101 on the CIFAR-100 dataset.}
    \begin{tabular}{lcc}
      \hline
      \textbf{Backbone}        & \textbf{Fixed} & \textbf{Ours} \\ \hline
      WideRes28$\times$2       & 73.83          & \textbf{75.61} \\
      ResNet-50                & 78.49          & \textbf{81.19} \\
      ResNet-101               & 80.97          & \textbf{83.45} \\ \hline
    \end{tabular}
    \label{tab:multi_backbone}
\end{table}

 \subsection{Comparison with Hierarchical-aware Prototype Positioning }
We conducted experiments comparing our method with hierarchical-aware prototype positioning, i.e., Hyperbolic Bussman Learning (HBL) \citep{ghadimi2021hyperbolic}. Specifically, we replaced the backbone in \citep{ghadimi2021hyperbolic} with the same backbone (Wide ResNet-28-2) used in our method, and set the dimension to 128, consistent with our approach. The results shown in ~\Cref{tab:bussman} indicate that our method still exhibits significant superiority. We will add these results in the revised version.

\begin{table}[hbp]
  \centering
  \small
    \caption{Comparison with Hyperbolic Bussman Learning (HBL)~\citep{ghadimi2021hyperbolic} on the CIFAR-10 and CIFAR-100 datasets. * indicates results obtained by our own implementation.}
    \begin{tabular}{lcc}
      \hline
      \textbf{Method} & \textbf{CIFAR-10} (\%) & \textbf{CIFAR-100} (\%) \\ \hline
      HBL*            & 91.16                 & 67.42                  \\
      Ours            & \textbf{94.75}        & \textbf{75.61}         \\ \hline
    \end{tabular}
    \label{tab:bussman}
\end{table}

 \begin{figure*}[htp]
\centering
\includegraphics[width=0.6\linewidth]{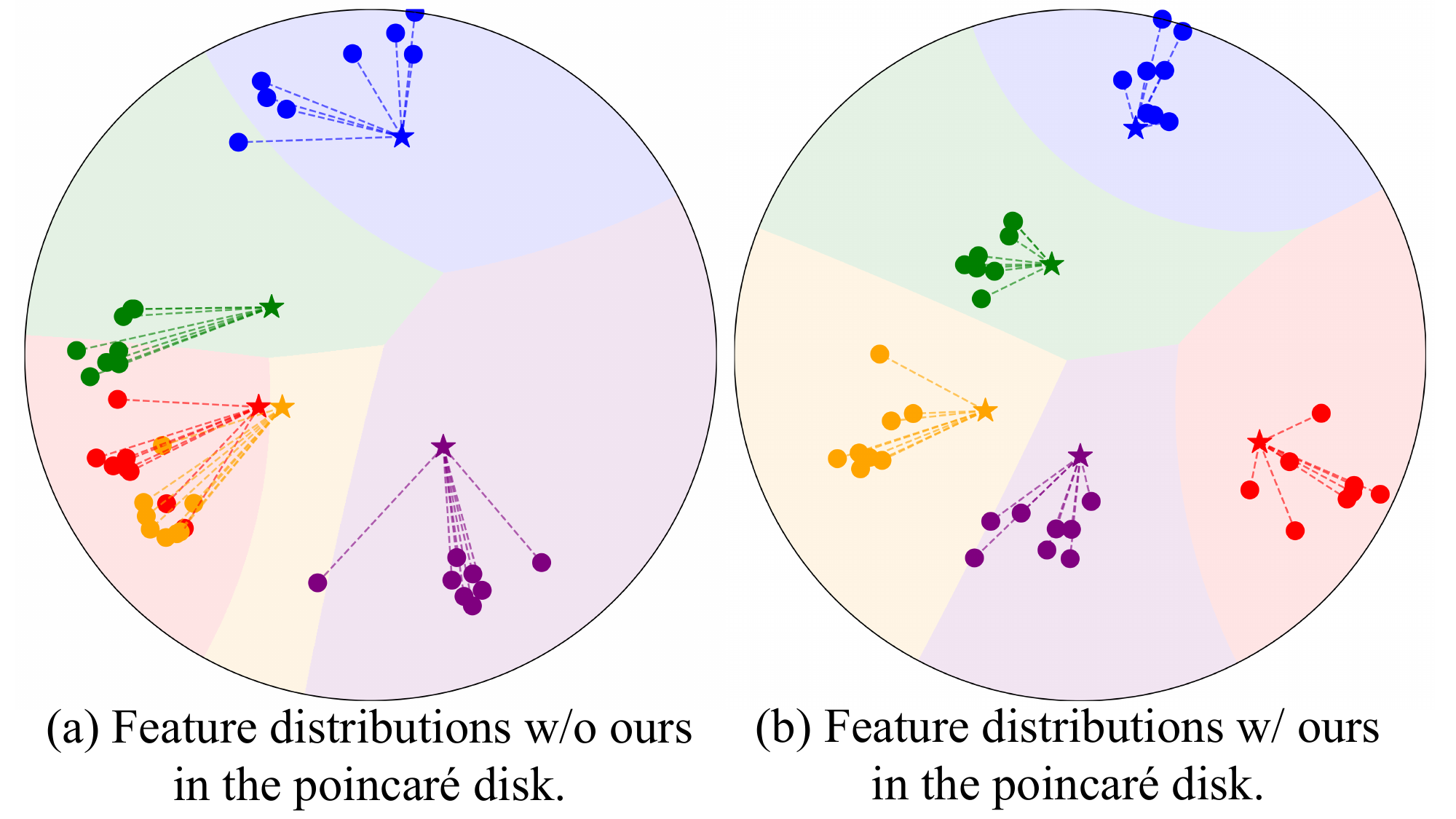}

\caption{Feature distribution on the mini-ImageNet dataset for 5-ways, 5-shots setting with 8 query samples. Dotted lines connect prototypes ($\star$) and query samples ($\bullet$). Shaded regions represent classification areas, with colors indicating categories. Comparison is shown for w/o and w/ our method. }
\label{fig:vis}
 
\end{figure*}

\subsection{Visualization}
\subsubsection{Hyperbolic Embedding Distribution}
We present estimated feature distributions for the mini-ImageNet dataset using horopca~\citep{chami2021horopca} for dimensionality reduction. We apply the 1-nearest-neighbor algorithm (with Poincaré distance) to compute classification boundaries, shown in \Cref{fig:vis}. Our method (\Cref{fig:vis}(b)) corrects misclassifications presented in \Cref{fig:vis}(a), resulting in more uniform classification zones. The distance between query points and prototypes is closer, improving cohesion within categories. For example, in \Cref{fig:vis}(a), yellow and red prototypes are closely positioned, causing yellow query samples to fall into the red zone. Our method (\Cref{fig:vis}(b)) effectively separates these prototypes, correcting the misclassification. By expanding distances between prototypes, our method enhances discriminative capabilities and strengthens class identification by clustering query samples near the new prototypes, leading to clearer class separation.

\begin{figure*}[htbp]
    \centering
    \includegraphics[width=0.8\linewidth]{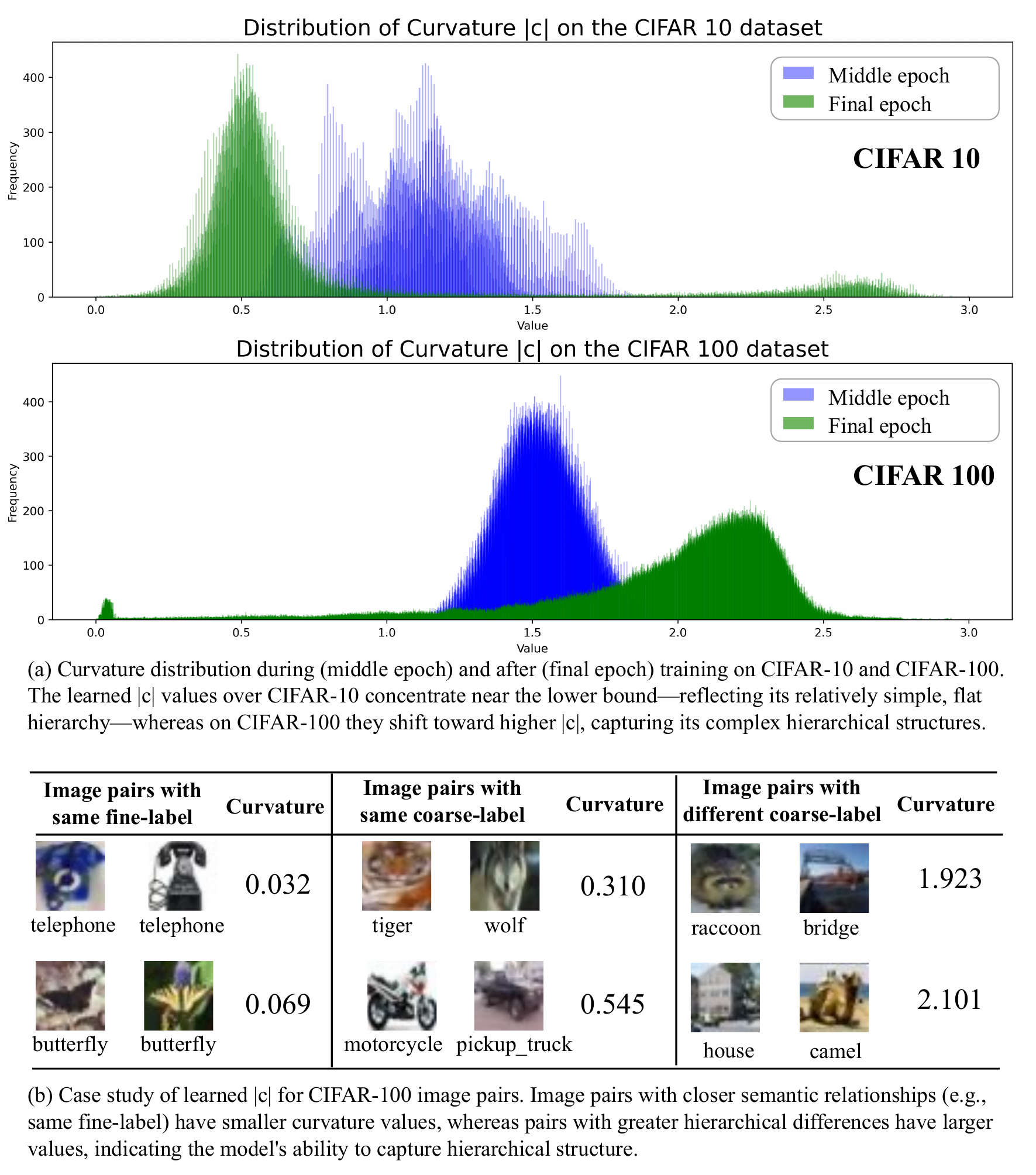}
    \caption{Curvature distribution during (middle epoch) and after (final epoch) training and case study. The range of the learned
curvature is set to [0.0001, 3.0] in the experiment of (a). In (b), image pairs with closer levels have lower curvature $|c|$, while
image pairs with greater hierarchical differences have larger $|c|$.} 
    \label{fig:cur}
\end{figure*}
 
\subsubsection{Curvature Distribution}
We visualize the curvature distributions in the middle epoch and the final epoch when training on the CIFAR-10 and CIFAR-100 datasets, as shown in \Cref{fig:cur}(a). We observe that the training process in CIFAR-10 pushes the curvatures to small values, while the training process in CIFAR-100 pushes the curvatures to large values. The reason is that the CIFAR-10 dataset has a relatively simple hierarchical structure, while CIFAR-100 has a complex hierarchical structure. This is also confirmed by the delta hyperbolicity values in \Cref{tab:delta}, where CIFAR-100 has a smaller delta hyperbolicity value than CIFAR-10, showing the more complex hierarchical structure in CIFAR-100.

We also conduct a case study about the curvatures and hierarchical relationships in CIFAR-100 image pairs. As shown in \Cref{fig:cur} (b), image pairs with closer levels are assigned lower curvature values, while those with greater hierarchical differences receive larger $|c|$ values. Our method tends to assign lower curvature to semantically closer image pairs, highlighting its effectiveness in capturing hierarchical differences in image representations.
 
\subsubsection{Hard Cases}
\begin{figure*}
    \centering
    \includegraphics[width=1\linewidth]{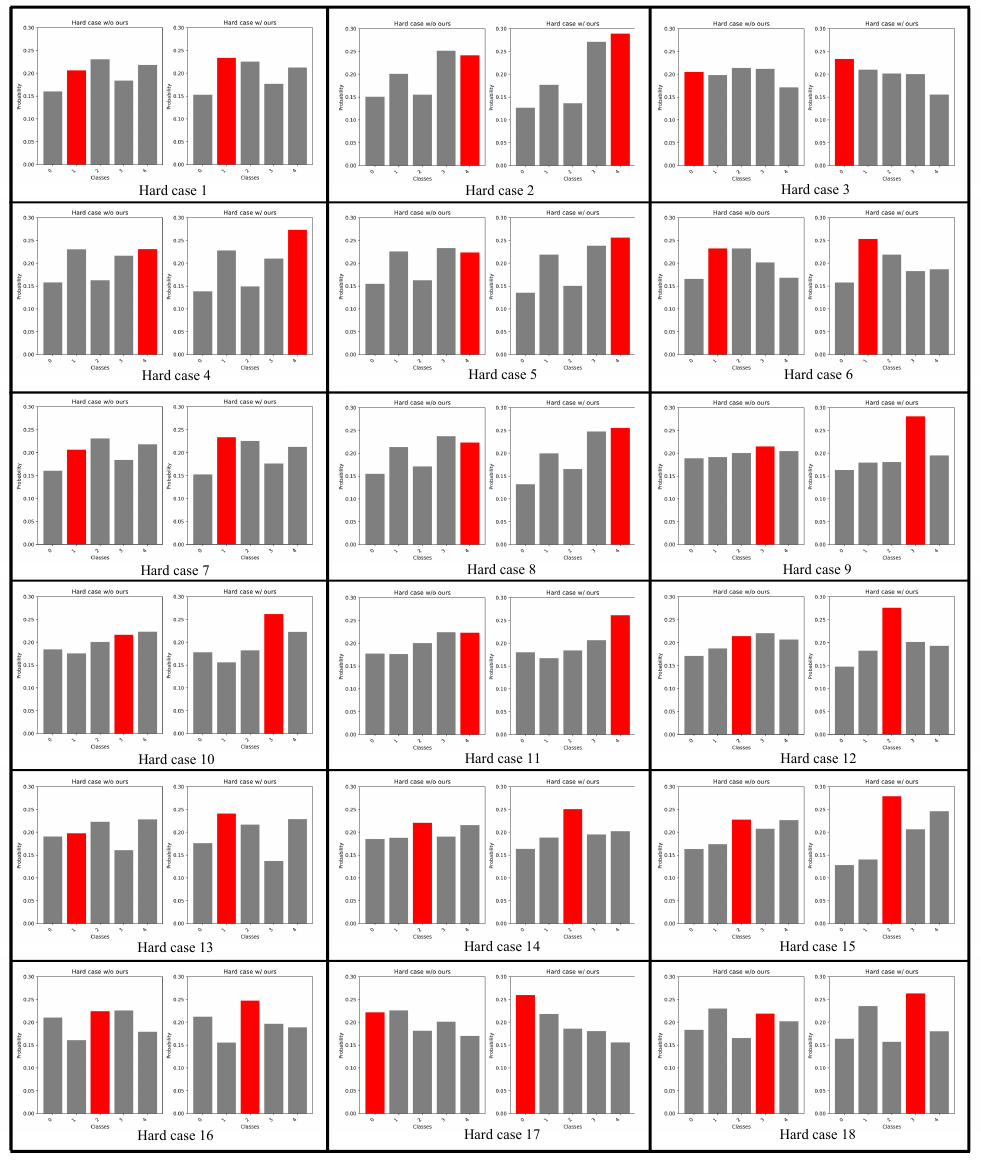}
    \caption{Probability distributions of several hard cases in 5-way 5-shot classification on the mini-ImageNet dataset. In each cell, the left histogram shows predictions without our method, and the right shows predictions after applying our method. The red bar indicates the correct class. After applying our method, the correct class receives higher probability, showing that our approach effectively reduces misclassification in challenging cases by refining the distance-based decision boundaries.}
    \label{fig:hard-dis-example}
\end{figure*}

We visually compare the logits before and after applying our method to hard cases. We calculate the
probabilities by applying the softmax function to the logits, and the results are shown in \Cref{fig:hard-dis-example}.
Each cell in \Cref{fig:hard-dis-example} represents a query sample’s probabilities of 5-ways task, where each bar denotes
the probability of being classified into a specific category, with the red bar indicating the correct
category. Within each cell of the \Cref{fig:hard-dis-example}, the left figure represents the probabilities without our
method, and the right figure demonstrates that using our method. It can be seen that, after applying
our method to hard cases, the probability of the correct class increases, indicating that
our method is able to correct misclassification caused by fixed distance measures.

\section{Conclusion}

In this paper, we have presented geometry-aware hyperbolic distance measures that accommodate diverse hierarchical data structures through adaptive projection matrix and curvature.
The adaptive curvature endows embeddings with more flexible hyperbolic spaces that better match the inherent hierarchical structures.
The low-rank projection matrices bring positive pairs closer and push negative pairs farther apart.
Moreover, the hard-pair mining mechanism enables the efficient selection of hard cases from the query set without introducing additional parameters, reducing the computational cost. 
Theoretical analysis and experiments show the effectiveness of our method in refining hyperbolic learning through geometry-aware distance measures.
In the future, we will explore employing our method in the multi-modal setting and large language models, extending the application of hyperbolic learning algorithms.


\bibliography{main}

\end{document}